\PassOptionsToPackage{table}{xcolor}
\documentclass[10pt,twocolumn,letterpaper]{article}

\usepackage{iccv}              

\usepackage{colortbl}
\usepackage{multirow}
\usepackage{amssymb}
\usepackage{booktabs}
\usepackage{tabularx}
\usepackage{graphicx} 
\usepackage{comment}
\usepackage{url}

\definecolor{deepPurple}{RGB}{162, 0, 112}
\definecolor{maroon}{RGB}{188, 76, 0}
\definecolor{deepYellow}{RGB}{255, 192, 0.0}
\definecolor{lightBlue}{RGB}{20,134,180}
\definecolor{deepGreen}{RGB}{99, 141, 31}  
\definecolor{lightGreen}{RGB}{21, 155, 123}  
\definecolor{deepRed}{RGB}{192, 0, 0}

\definecolor{iccvblue}{rgb}{0.21,0.49,0.74}
\usepackage[pagebackref,breaklinks,colorlinks,allcolors=iccvblue]{hyperref}

\def\confName{ICCV}
\def\confYear{2025}

\title{VRU-Accident: A Vision-Language Benchmark for Video Question Answering and Dense Captioning for Accident Scene Understanding}

\author{Younggun Kim \hspace{1.5em}
Ahmed S. Abdelrahman\thanks{* Corresponding Author.} \hspace{1.5em}
Mohamed Abdel-Aty\\
University of Central Florida, United States\\
{\tt\small \{younggun.kim, ahmed.abdelrahman, m.aty\}@ucf.edu} \\
}

\begin{document}

\twocolumn[{%
  \maketitle
  \begin{center}
      \vspace{-3em}
      {\color{magenta}
      \textbf{https://vru-accident.github.io/}}

  \end{center}
}]

\renewcommand*{\thefootnote}{\fnsymbol{footnote}}
\footnotetext[1]{Corresponding Author.}

\begin{abstract}
Ensuring the safety of vulnerable road users (VRUs), such as pedestrians and cyclists, is a critical challenge for autonomous driving systems, as crashes involving VRUs often result in severe or fatal consequences. While multimodal large language models (MLLMs) have shown promise in enhancing scene understanding and decision-making in autonomous vehicles, there is currently no standardized benchmark to quantitatively evaluate their reasoning abilities in complex, safety-critical scenarios involving VRUs. To address this gap, we present \textbf{VRU-Accident}, a large-scale vision-language benchmark designed to evaluate MLLMs in high-risk traffic scenarios involving VRUs. VRU-Accident comprises \textbf{1K} real-world dashcam accident videos, annotated with \textbf{6K} multiple-choice question-answer pairs across six safety-critical categories (with \textbf{24K} candidate options and \textbf{3.4K} unique answer choices), as well as \textbf{1K} dense scene descriptions. Unlike prior works, our benchmark focuses explicitly on VRU-vehicle accidents, providing rich, fine-grained annotations that capture both spatial-temporal dynamics and causal semantics of accidents. To assess the current landscape of MLLMs, we conduct a comprehensive evaluation of \textbf{17} state-of-the-art models on the multiple-choice VQA task and on the dense captioning task. Our findings reveal that while MLLMs perform reasonably well on visually grounded attributes, they face significant challenges in reasoning and describing accident causes, types, and preventability. 
\end{abstract}

\section{Introduction}
\label{sec:intro}

\begin{figure}[t!]
\centering
\includegraphics[width=\columnwidth, trim=130 20 130 20, clip]{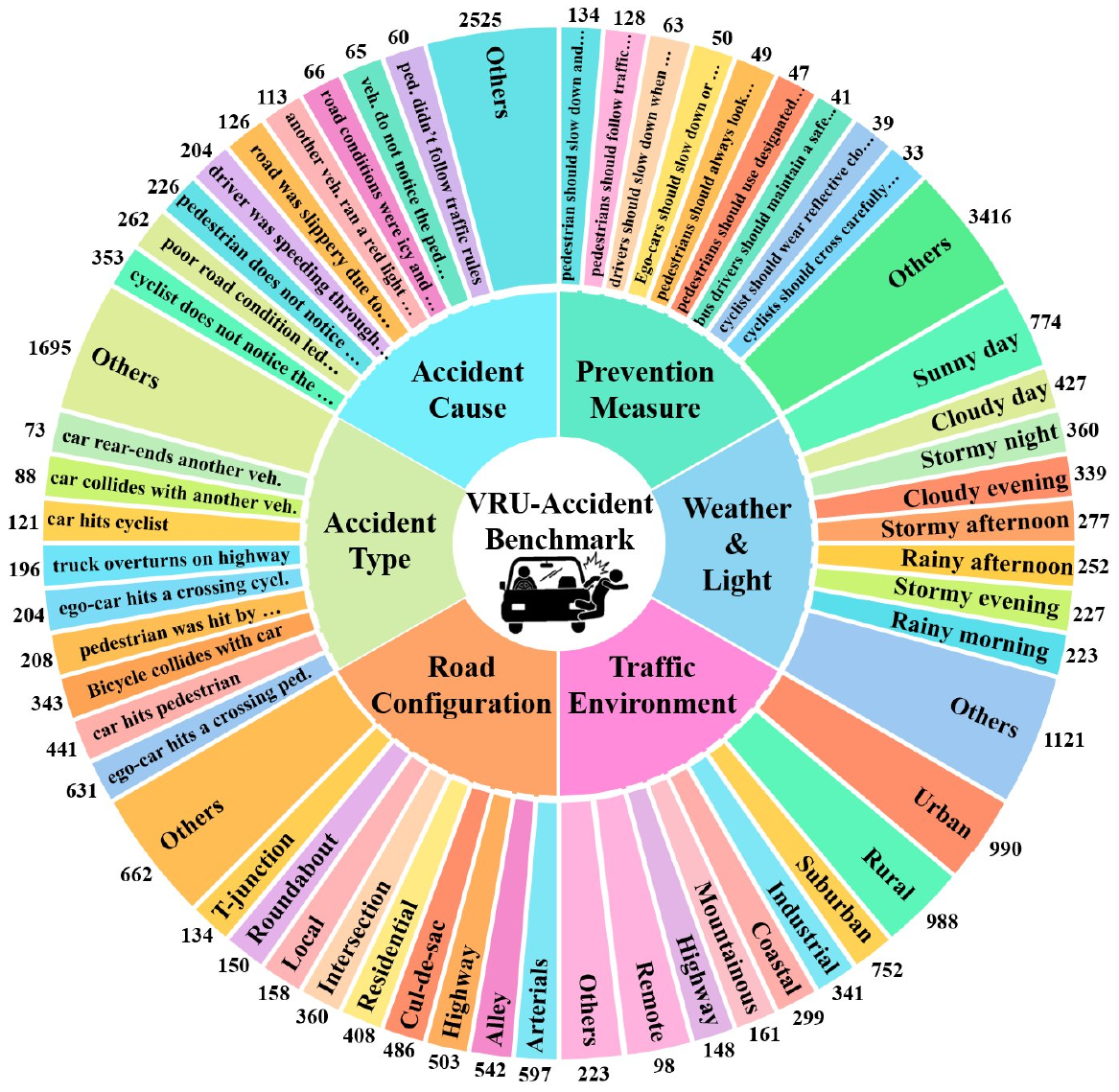}
\vspace{-1em}
\caption{VQA distribution in the VRU-Accident benchmark.}
\label{fig:distribution}
\end{figure}

Multimodal Large Language Models (MLLMs) have become an important research area, showing strong performance in video understanding tasks such as video captioning, visual question answering, and visual grounding~\cite{MLLM2, you2023ferret, abdelrahman2025video}. MLLMs have exhibited robust generalization across diverse downstream tasks, attributed to pretraining on large-scale, heterogeneous multimodal corpora~\cite{MLLM1, MLLM3, MLLM5}. This generalizability has sparked increasing interest in applying MLLMs to many scene understanding tasks which is critical to deepen our understanding of machine understanding and reasoning. To facilitate such applications, researchers have developed benchmarks and datasets focused on traffic event-related question-answering and general scene understanding~\cite{TrafficQA1, TrafficQA2, abdelrahman2025advanced}. However, these datasets overlook accident-centric reasoning, which is critical for safe driving perception~\cite{ham2023cipf, abdelrahman2025vru}. 

Recently, several accident datasets~\cite{CTA, DADA_2000, CAP_DATA, MM_AU} provide short casual captions per video that describe accident types, causes, and possible prevention strategies, providing context for accident interpretation. Despite their utility, these video-based vision-language accident datasets rely on human annotators to classify accidents into predefined categories, which restricts their ability to evaluate the generalization capabilities of MLLMs on accident scenarios, unlike video question-answer (VQA) settings that require models to reason over dynamically constructed and semantically diverse answer spaces. Moreover, these datasets lack detailed accident scene descriptions, making it difficult to assess an MLLM's ability to generate context-rich and semantically accurate narratives about accident events. Finally, these datasets are heavily biased toward vehicle-to-vehicle collisions, resulting in under-representation of incidents involving VRUs, such as pedestrians and cyclists, where they VRUs accidents are more dangerous and lead to fatalities~\cite{abdelrahman2025vrucrosssafe, ahsan2025evaluating, VRU1}.

VRUs are among the most at-risk road users in traffic environments, where in 2020 alone, 6,516 pedestrians and 938 bicyclists lost their lives in traffic accidentes with 3.9\% and 9\% increase from 2019, respectively, and over 54,000 were injured~\cite{abdelrahman2025vrucrosssafe}. Alarmingly, pedestrian deaths accounted for 17\% of all traffic fatalities, reflecting a 53\% increase since 2009~\cite{GHSA2020}. Despite the severity and frequency of such incidents, we lack extensive vision-language benchmark specifically designed to evaluate the capabilities of MLLMs in answering questions or generating descriptions related to VRU-involved accidents with vehicles. This absence hinders the development and assessment of MLLMs in safety-critical applications such as autonomous vehicles, where evaluating a model’s ability to understand and reason about VRU-related accident scenarios is essential and ensure their driving perception for safe navigation even under adverse weather conditions.

To address this gap, we propose \textbf{VRU-Accident}, a large-scale benchmark designed to evaluate the reasoning and description capabilities of MLLMs in VRU traffic accidents. The benchmark includes 1,000 real-world accident and near-accident videos annotated with diverse VQA pairs and dense scene descriptions, covering various contextual and causal aspects of each incident. It serves as the first benchmark specifically curated to assess both VQA and captioning performance for safety-critical scenarios involving VRUs, where Figure~\ref{fig:distribution} shows the distribution of the VQA task of each category in VRU-Accident benchmark. 
Our key contributions are summarized as follows:
\begin{itemize}
    \item We introduce VRU-Accident, a large-scale benchmark comprising \textbf{1K} VRU-related accident videos, \textbf{6K} VQA questions with \textbf{24K} candidate options, and \textbf{1K} dense scene-level captions.
    
    \item We propose a semi-automatic benchmark curation pipeline that generates diverse and semantically rich VQA candidate sets. Each question includes one correct answer and three contextually plausible counterfactual, resulting in over \textbf{3.4K} unique answer options and making the task challenging for MLLMs.
    
    \item We conduct extensive evaluations of \textbf{17} MLLMs, with 15 open-source and 2 closed-source, on the VQA task and the dense captioning task, providing a unified benchmark for assessing their performance in safety-critical scene understanding task while offering key insights into their limitations.
    
\end{itemize}

\section{Related Work}
\label{sec:relatedwork}

\paragraph{Video QA Datasets and Benchmarks in Transportation Domain}
Recent advances in vision-language models have spurred the development of various VQA datasets and benchmarks within the traffic domain. Several recent works~\cite{TrafficQA1, TrafficQA2, TrafficQA3, TrafficQA4} have proposed multi-modal VQA benchmarks in the context of autonomous driving, aiming to evaluate high-level reasoning about road scenes. Another research~\cite{TrafficQA10, TrafficQA11, TrafficQA8} investigates language grounding and explanation in complex traffic scenes. These datasets focus on providing textual rationales or identifying key objects and behaviors, thereby enhancing interpretability and supporting downstream decision making. The other works~\cite{TrafficQA7, TrafficQA9} introduce task-specific benchmarks that target action recognition, risk localization, and importance-based reasoning. These efforts promote temporally grounded event understanding and knowledge extraction via joint captioning and QA. Despite their success in advancing scene understanding and reasoning in traffic environments, these benchmarks do not specifically address accident scenarios, particularly those involving vulnerable road users (VRUs) such as pedestrians and cyclists. As a result, they offer limited utility for evaluating a model’s capacity to reason about accident dynamics, causality, and preventability in safety-critical contexts where accurate accident comprehension is essential.


\begin{table}[!t] 
\centering
\caption{Comparison of accident video datasets. Surv.: Surveillance View, VQA: Video Question Answering, CC: Casual Captioning, DC: Dense Captioning, and R/S: Real/Synthetic videos.}
\label{tab:AccidentDataset}
\resizebox{\columnwidth}{!}{%
\begin{tabular}{lcccccccc}
\toprule
\textbf{Datasets} & \textbf{View} & \textbf{\#Clips} & \textbf{\#VRU accident} & \textbf{VQA} & \textbf{CC} & \textbf{DC} & \textbf{R/S} \\
\midrule

ROL ~\cite{ROL} & Dashcam & 1,000 & - & & & &  R \\

DeepAccident ~\cite{DeepAccident} & Dashcam & - & - & & & & S \\

CTA ~\cite{CTA} & Dashcam & 1,935 & - & & \checkmark &  & R \\

CTAD ~\cite{CTAD} & Surv. & 1,100 & - & &  &  & S \\

SUTD-TrafficQA ~\cite{SUTD_TrafficQA} & Surv. & 10,080 & - & \checkmark & & & R \\

TUMTraffic-VideoQA ~\cite{TUMTraf-VideoQA} & Surv. & 1,000 & $<$100 & \checkmark & \checkmark & \checkmark & R \\

TUMTraf-A ~\cite{TUMTraf-A} & Surv. & 48 & - &  &  &  & R \\

A3D ~\cite{A3D} & Dashcam & 3,757 & $<$100 &  & & & R \\

DoTA ~\cite{DoTA} & Dashcam & 5,586 & 100 &  & & & R \\

DADA-2000 ~\cite{DADA_2000} & Dashcam & 2,000 & 223 & & \checkmark & & R \\

MM-AU ~\cite{MM_AU, CAP_DATA} & Dashcam & \textbf{11,727} & 510 & & \checkmark & & R \\
\hline
\textbf{VRU-Accident (Ours)} & Dashcam & 1,000 & \textbf{1,000} & \checkmark & \checkmark & \checkmark & R \\

\bottomrule
\end{tabular}
}
\end{table}

\begin{table*}[t]
\centering
\vspace{-0.8em}
\caption{Comparison of VRU-related accident distributions across accident datasets, DoTA, DADA-2000, MM-AU, and our proposed VRU-Accident benchmark. VRU-Accident provides broader coverage across all attributes.}
\label{tab:RelatedDatasetDistribution}
\resizebox{\textwidth}{!}{%
\begin{tabular}{l|cc|cc|cccc|ccccc|cccccc}
\toprule
\multirow{2}{*}{\textbf{Datasets}}& \multicolumn{2}{c|}{\textbf{VRU Types}} & \multicolumn{2}{c|}{\textbf{Lightning}} & \multicolumn{4}{c|}{\textbf{Weather Condition}} & \multicolumn{5}{c}{\textbf{Road Configuration}}  & \multicolumn{6}{|c}{\textbf{Traffic Environment}} \\
\cline{2-20}
 & pedestrian & cyclist & day & night & sunny & rainy & snowy & cloudy & arterials & intersection & t-junction & curve  & others & highway & urban & suburban & rural & mountain & tunnel \\
\midrule

DoTA ~\cite{DoTA} & 89 & 11 & 79 & 21 & 86 & 2 & 12 & 0 & 75 & 14 & 10 & 1 & 0 &  2 & 89 & 0 & 9 & 0 &  0 \\

DADA-2000 ~\cite{DADA_2000} & 114 & 109 &  197 & 26 & 213 & 10 & 0 & 0 & 84 & 87 & 43 & 7 & 2 & 1 & 186 & 0 & 33 & 2 & \textbf{1} \\

MM-AU ~\cite{MM_AU, CAP_DATA} & 320 & 190 & 452 & 58 & 418 & 29 & 46 & 17 & 248 & 163 & 71 & 16 & 12 &  3 & 430 & \textbf{6} & 68 & 2 & \textbf{1} \\

\hline
\textbf{VRU-Accident(Ours)} & \textbf{730} & \textbf{270} & \textbf{873} & \textbf{127} & \textbf{808} & \textbf{61} & \textbf{108} & \textbf{23} & \textbf{584} & \textbf{241} & \textbf{133} & \textbf{24} & \textbf{18} & \textbf{6} & \textbf{873} & \textbf{6} & \textbf{111} & \textbf{3} &\textbf{1}  \\
\bottomrule
\end{tabular}
}
\end{table*}

\begin{figure*}[!t]
    \centerline{\includegraphics[width=\textwidth]{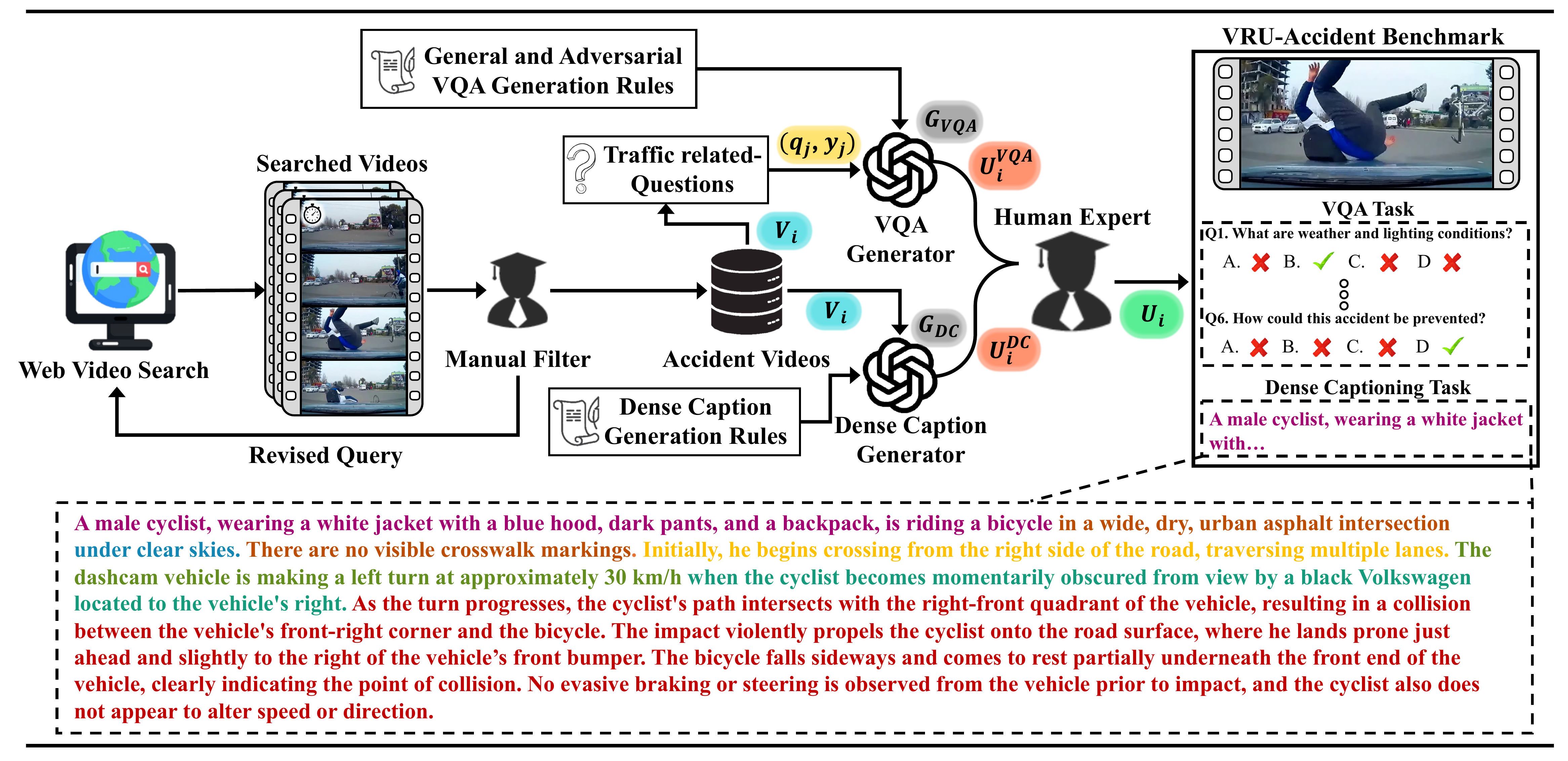}}
    \vspace{-1em}
    \caption{Overview of the VRU-Accident benchmark curation pipeline. Accident videos are retrieved from web sources based on traffic-related queries and manually filtered by human experts. For each video $V_i$, a set of ground truth answers $y_j$ is first annotated by human experts for each question $q_j$. Then, the VQA generator $G_{\text{VQA}}$ generates three counterfactual answers $y_j^*$ based on $(q_j, y_j)$. Separately, a dense accident description $C_i$ is generated for $V_i$ using the dense captioning generator $G_{\text{DC}}$. All annotations are verified by human experts before being finalized. Colors: \textcolor{deepPurple}{vehicle or pedestrian appearances}, \textcolor{maroon}{environmental factors}, \textcolor{lightBlue}{weather conditions}, \textcolor{deepYellow}{pedestrian's kinematic features}, \textcolor{deepGreen}{vehicle's kinematic features}, \textcolor{lightGreen}{spatial relationship between road users}, and \textcolor{deepRed}{detailed collision descriptions}.}
    \label{fig:benchmark_pipeline} 
\end{figure*}

\paragraph{Accident Understanding Datasets and Benchmarks}

To support research on accident understanding in driving scenarios, several video datasets have been proposed in recent years. A summary of representative accident-centric datasets is presented in Table~\ref{tab:AccidentDataset}, while Table~\ref{tab:RelatedDatasetDistribution} provides detailed distributions for the subset of ego-view datasets that contain VRU involved accident videos. These datasets vary in terms of view type, volume, annotation modalities, and coverage of accident content. ROL~\cite{ROL}, A3D~\cite{A3D}, and DoTA~\cite{DoTA} focus on anomaly detection in dashcam videos, while CTAD~\cite{CTAD} and DeepAccident~\cite{DeepAccident} leverage synthetic or simulation-based video generation to model urban traffic accidents. SUTD-TrafficQA~\cite{SUTD_TrafficQA} is a benchmark designed for VQA tasks in traffic events but relies on surveillance footage. In contrast, DADA-2000~\cite{DADA_2000} includes driver attention data to study accident prediction, and MM-AU~\cite{MM_AU}, which among all their videos only 510 are accidents involving VRUs, with 223 videos already imported from the DADA-2000 dataset, present large-scale collections of real-world dashcam incidents for visual reasoning and forecasting. Among them, a few works~\cite{CTA, DADA_2000, CAP_DATA, MM_AU} include short captions describing accident types, causes, or prevention strategies. However, these captions are entirely dependent on human annotators and constrained to predefined categories, which not only limits their expressiveness in capturing the diverse semantics of real-world accident scenarios but also restricts the ability to evaluate the generalization capabilities of modern MLLMs such as LLaVA family~\cite{LLaVA_OneVision, LLaVA_NeXT_Video, LLaVA_Video}, InternVL family~\cite{InternVL2, InternVL25, InternVL3}, Qwen family~\cite{qwen2_VL, qwen25_VL}, and the others~\cite{Mobile_VideoGPT, Video_XL_Pro, Video_XL2, GPT, gemini}. One the other hand, SUTD-TrafficQA~\cite{SUTD_TrafficQA} addresses this gap by introducing a structured VQA benchmark for traffic scenarios, enabling more principled evaluation of VQA performance in MLLMs. However, its reliance on surveillance camera footage limits its ability to capture direct interactions between vehicles and VRUs, such as pedestrians and cyclists, especially during moments of impact or collision. Our VRU-Accident benchmark is set to target VRU-related accidents in a vision-language framework. It combines the MM-AU and DoTA samples while adding more samples reaching 1,000 videos. It includes VQA annotations across six accident-related categories and detailed dense captions for 1,000 real-world ego-view accident videos. This enables a comprehensive and fine-grained evaluation of MLLMs’ reasoning and description capabilities in realistic, high-risk interaction scenarios between vehicles and VRUs.


\section{VRU-Accident Benchmark}
\label{sec:dataset_curation}

In this section, we describe the overall design of the VRU-Accident benchmark, focusing on two key components: the benchmark curation process and statistics. Our goal is to establish a standardized and comprehensive framework for analyzing accident scenes involving VRUs through both VQA and Dense Captioning tasks. We aim to capture a wide range of contextual, environmental, and causal elements from real-world traffic accident videos to facilitate multimodal reasoning in safety-critical scenarios. We focus on collecting real-world traffic accident videos involving VRUs and curating a benchmark tailored for two vision-language tasks: VQA and Dense Captioning. The overall curation pipeline is illustrated in Figure~\ref{fig:benchmark_pipeline}. We define each accident video as $V_i$, where $i = 1, 2, \ldots, N$.

\begin{figure*}[!h]
    \centerline{
        \includegraphics[width=0.97\textwidth, trim=100 15 100 15, clip]{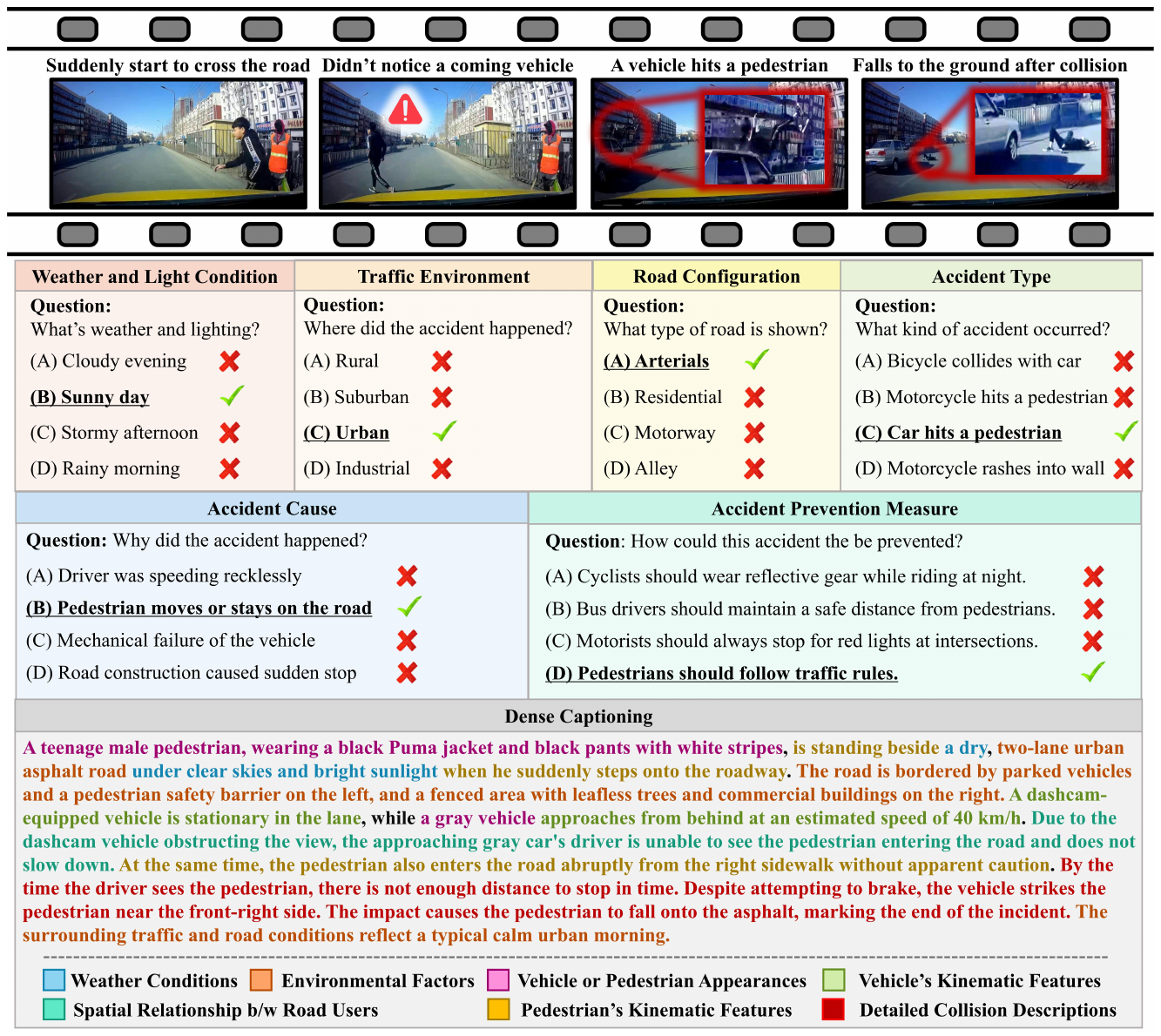}
    }
    \caption{Visualization of annotations in VRU-Accident benchmark illustrating both the VQA and dense captioning tasks.}
    \label{fig:VQA_Caption} 
\end{figure*}

\subsection{VQA Category Selection}
We define six core VQA categories as $j \in$ \{1. \textit{Weather \& Light}, 2. \textit{Traffic Environment}, 3. \textit{Road Configuration}, 4. \textit{Accident Type}, 5. \textit{Accident Cause}, 6. \textit{Accident Prevention Measure}\}. Among these, \textit{Accident Type}, \textit{Accident Cause}, and \textit{Accident Prevention Measure} are essential for understanding the core dynamics and outcomes of VRU-involved accidents, and have been widely used in prior research on traffic incident analysis ~\cite{MM_AU, SUTD_TrafficQA, DADA_2000, CTA}. In addition, we include contextual attributes such as \textit{Weather \& Light}, \textit{Traffic Environment}, and \textit{Road Configuration}, which are known to significantly influence accident circumstances and situational awareness ~\cite{Lei1, Lei2, Lei3, Wang1}.

\subsection{VQA Curation Pipeline}
Our task is framed as a multi-category VQA problem. For each video $V_i$, we define a set of questions and corresponding answers as $Q_i$ and $Y_i$, respectively. Each $Q_i$ and $Y_i$ contain a question-answer pair $(q_j, y_j)$ for each of the six VQA categories, where $j = 1, \dots, 6$. To enable discriminative evaluation, we additionally define a set of three counterfactual answers $y_j^*$ per category, which belong to the same domain as $y_j$ but are incorrect. These are generated using a VQA counterfactual generation function $G_{\text{VQA}}$, such that $y_j^* = G_{\text{VQA}}(q_j, y_j)$. We then define the final candidate answer set for each question as $\hat{Y}_j = \{y_j, y_j^{*(1)}, y_j^{*(2)}, y_j^{*(3)}\}$. Each video's VQA annotation is summarized as $\mathcal{U}^{VQA}_{i} = \{V_i, Q_i, \hat{Y}_i\}$, where $\hat{Y}_i = \{\hat{Y}_j\}_{j=1}^6$ denotes the candidate answer sets for all six categories.

\subsection{Dense Caption Curation Pipeline} The dense captioning task aims to generate a comprehensive and temporally grounded textual description of the traffic accident scenario depicted in each video $V_i$. Rather than enumerating every visual element, we focus on capturing the essential dynamics of the incident, such as weather and environmental factors, road users' appearance and posture, road users' kinematic features, and the sequential interactions between the vehicle and the VRU leading up to the accident. The descriptions are required to be concise yet information-dense, suitable for use as high-quality ground truth data in accident analysis. To achieve this, we employ a Dense Captioning generator $G_{\text{DC}}$ that produces a caption $C_i$ for each video $V_i$, such that $C_i = G_{\text{DC}}(V_i)$. The final annotation for the dense captioning task is represented as $\mathcal{U}^{\text{DC}}_i = \{V_i, C_i\}$.
Finally, the complete annotation for each video of VRU-Accident is defined as $\mathcal{U}_i = \mathcal{U}^{\text{VQA}}_i \cup \mathcal{U}^{\text{DC}}_i = \{V_i, Q_i, \hat{Y}_i, C_i\}$, and Figure~\ref{fig:VQA_Caption} illustrates an example of both VQA and Dense Captioning annotations in our VRU-Accident benchmark.

 \begin{table}[!t] 
\centering
\vspace{-1.2em}
\caption{Summary of annotation statistics in the VRU-Accident benchmark.}
  \label{table:VRU_Accident}
    {\resizebox{\columnwidth}{!}{
\begin{tabular}{c|c||c|c|c}
\hline

Task & Categories & Options/GT  &   \# Samples & \# Unique Samples \\
    \hline \hline

\multirow{12}{*}{VQA} & \multirow{2}{*}{Weather \& Light} & Options& 4K & 38 \\
\cline{3-5}
 & & Ground Truth& 1K  & 13 \\
\cline{2-5}

& \multirow{2}{*}{Traffic Environment} & Options& 4K  & 27 \\
\cline{3-5}
 & & Ground Truth&  1K & 6 \\
\cline{2-5}
    
& \multirow{2}{*}{Road Configiguration} & Options&  4K & 42 \\
\cline{3-5}
& & Ground Truth&  1K & 10 \\
\cline{2-5}
    
& \multirow{2}{*}{Accident Type} & Options&  4K & 405 \\
\cline{3-5}
& & Ground Truth& 1K  & 33 \\
\cline{2-5}
    
& \multirow{2}{*}{Accident Cause} & Options&  4K & 782 \\
\cline{3-5}
& & Ground Truth& 1K  & 135 \\
\cline{2-5}
    
& \multirow{2}{*}{Prevention Measure} & Options&  4K & 2143 \\
\cline{3-5}
& & Ground Truth&  1K & 211 \\
 
\hline
\hline
Dense & \multicolumn{2}{c|}{\multirow{2}{*}{ Detailed Description of Accident Video}} & \multirow{2}{*}{1K} & \multirow{2}{*}{1K} \\
Captioning & \multicolumn{2}{c|}{} &  &  \\
\hline

\end{tabular}
}}
\end{table}

\subsection{Benchmark Statistics}
The VRU-Accident benchmark contains a total of 1,000 real-world traffic accident videos of VRUs. Each video is annotated with both VQA and dense captioning labels. As shown in Table~\ref{table:VRU_Accident}, the VQA task comprises 6,000 question-answer (QA) pairs and 1,000 dense captioning. For each QA pair, we construct a set of four candidate answers (one ground truth and three counterfactual options), resulting in a total of 24,000 multiple-choice options across the dataset. In addition to the volume of annotations, a key characteristic of VRU-Accident is the large number of unique answer samples per category. For example, the \textit{Prevention Measure} and \textit{Accident Cause} categories contain 2,143 and 782 unique textual answers, respectively. Unlikely accident classification tasks \cite{MM_AU, DADA_2000, DoTA} which require predefined number of classes in each category, high variability of our benchmark requires models not only to classify from a fixed label space but to reason about fine-grained and context-dependent semantics, underscoring the importance of VLMs for this benchmark. The dense captioning task also contributes to the benchmark’s complexity and richness, providing 1,000 detailed accident descriptions, resulting in 1,000 unique captions aligned with real-world accident scenarios. For more details about benchmark statistics, please refer to Supplementary Section~\ref{sec:metrics}.

\section{Experiments}
\label{sec:exp}

\begin{table*}[t]
\centering
\caption{Comparison of VQA performance of state-of-the-art models on VRU-Accident Benchmark. WL: Weather\&Light, TE: Traffic Environment, RC: Road Configuration, AT: Accident Type, AC: Accident Cause, and PM: Prevention Measure. Rows with gray color background represent closed-source models, while rows with white background denote open-source models. \textbf{Black}, \textbf{\textcolor{blue}{blue}}, and \textbf{\textcolor{red}{red}} colors indicate the best, second-best, and worst performance, respectively. \textbf{\underline{Underline}} denotes the best performance among open-source models.}
\label{tab:VQA}
\begin{tabularx}{\textwidth}{>{\hsize=1.5\hsize}l|l|*{7}{>{\centering\arraybackslash}X}}
\hline
Model (Param.) & Year & \textbf{$Acc_{WL}$ } & \textbf{$Acc_{TE}$ } & \textbf{$Acc_{RC}$ } & \textbf{$Acc_{AT}$ } & \textbf{$Acc_{AC}$ } & \textbf{$Acc_{PM}$ } & \textbf{$Acc_{AVG.}$} \\
\hline\hline
LLaVA-OneVision(0.5B) \cite{LLaVA_OneVision} & 2024 & \textbf{\textcolor{blue}{72.1}} & 79.9 & 40.1 & \textbf{\textcolor{red}{35.9}} & 34.8 & 40.8 & 50.6 \\

InternVL2.5(1B) \cite{InternVL25} & 2025 & 69.2 & 67.7 & 39.0 & 45.8 & \textbf{\textcolor{red}{21.3}} & 51.0 & 49.0 \\

Mobile-VideoGPT(1.5B) \cite{Mobile_VideoGPT} & 2025 & \textbf{\underline{78.4}} & 64.5 & 44.4 & 49.3 & 27.2 & 47.4 & 51.9 \\

InternVL2.5(2B) \cite{InternVL25} & 2025 & 68.3 & 79.6 & 46.8 & \textbf{\textcolor{blue}{\underline{70.7}}} & 38.0 & 50.3 & 59.0 \\

InternVL3(2B) \cite{InternVL3} & 2025 & 70.7 & 78.3 & 53.5 & 41.2 & \textbf{\textcolor{blue}{\underline{50.4}}} & 53.2 & 57.9 \\

Qwen2.5-VL(3B) \cite{qwen25_VL} & 2025 & 63.8 & 72.1 & 39.4 & 47.3 & 25.6 & 44.7 & \textbf{\textcolor{red}{48.8}} \\
Video-XL-Pro(3B) \cite{Video_XL_Pro} & 2025 & 69.0 & \textbf{\underline{87.5}} & 35.4 & 43.9 & 47.5 & 48.9 & 55.4 \\

InternVL2.5(4B) \cite{InternVL25} & 2025 & 69.9 & 82.7 & 42.9 & 56.7 & 50.1 & 51.0 & 58.9 \\

Video-XL2(7B) \cite{Video_XL2} & 2024 & 71.0 & 82.7 & 59.3 & 58.1 & 45.4 & 53.5 & 61.7 \\
LLaVA-NeXT-Video(7B) \cite{LLaVA_NeXT_Video} & 2024 & 71.6 & \textbf{\textcolor{blue}{84.1}} & 54.0 & 38.8 & 27.7 & \textbf{\textcolor{red}{33.1}} & 51.6 \\
LLaVA-Video (7B) \cite{LLaVA_Video} & 2024 & 68.2 & 80.7 & 64.7 & 63.7 & \textbf{\textcolor{blue}{\underline{50.4}}} & 58.6 & \textbf{\textcolor{blue}{\underline{64.4}}} \\

Qwen2-VL(7B) \cite{qwen2_VL} & 2024 & 70.3 & \textbf{\textcolor{red}{59.6}}  & \textbf{\textcolor{red}{34.1}} & 60.1 & 41.4 & 46.8 & 52.1 \\

InternVL2(8B) \cite{InternVL2} & 2024 & 61.9 & 78.1 & 42.8 & 58.3 & 43.6 & \textbf{\underline{62.0}} & 57.8 \\

InternVL2.5(8B) \cite{InternVL25} & 2025 & 67.6 & 81.9 & 51.4 & 48.0 & 48.1 & \textbf{\textcolor{blue}{61.9}} & 59.8 \\

InternVL3(8B) \cite{InternVL3} & 2025 & 70.0 & 81.8 & \textbf{\textcolor{blue}{\underline{67.8}}} & 64.8 & 43.7 & 58.4 & \textbf{\textcolor{blue}{\underline{64.4}}} \\

\hline

\rowcolor{gray!15}
GPT-4o-mini \cite{GPT} & 2024 & \textbf{\textcolor{red}{60.2}} & 78.6 & 42.1 & 46.3 & 34.8 & 50.3 & 52.1 \\
\rowcolor{gray!15}
Gemini 1.5-flash \cite{gemini} & 2024 & 65.7 & 78.5 & \textbf{71.4} & \textbf{77.9} & \textbf{54.6} & 53.0 & \textbf{66.9} \\
\hline
\rowcolor{green!10}
\textbf{Human Expert} & 2025 & \textbf{95.1} & \textbf{94.7} & \textbf{93.8} & \textbf{94.5} & \textbf{95.1} & \textbf{94.8} & \textbf{94.7}\\
\hline
\end{tabularx}
\end{table*}

\subsection{Implementation Details}

\textbf{Baseline MLLMs.} We evaluate a total of 17 SOTA MLLMs on the VRU-Accident benchmark. This includes 15 open-source models that are assessed on both the VQA and dense captioning tasks, and 2 closed-source models. All models are tested in a \textit{zero-shot} setting without any fine-tuning, ensuring a fair comparison of their generalization abilities in unseen VRU-related accident scenarios. All evaluations were conducted on a workstation equipped with 8 NVIDIA TITAN RTX GPUs, each with 24 GB memory.

\textbf{VQA Task.} We construct a total of \textbf{6K} VQA pairs based on \textbf{1K} VRU-related accident videos, with each video annotated with six question-answer pairs corresponding to distinct reasoning categories. Each question is accompanied by four candidate options (A, B, C, D), and models are prompted to select the correct choice. The generated responses are processed with model-specific postprocessing strategies to extract the predicted option, and \textit{category-wise accuracy} is computed to quantify performance.

\textbf{Dense Captioning Task.} We evaluate \textbf{1K} video-prompt pairs, where models are asked to generate a comprehensive description of each VRU-related accident video, including visual scene elements, road user behavior, vehicle dynamics, and accident events. Generated captions are evaluated using SPICE~\cite{SPICE}, METEOR~\cite{METEOR}, COMET~\cite{COMET}, and ROUGE~\cite{ROUGE} to assess linguistic quality and event coverage. BLEU score is not included, as it yields values below 0.1, due to its emphasis on exact n‑gram precision without incorporating recall or contextual semantics \cite{datta2022analysis}. We also present detailed evaluation metrics, full prompt templates, and implementation details for both VQA and Dense Captioning tasks in Supplementary Section~\ref{sec:metrics}.

\begin{figure}
\includegraphics[width=\columnwidth]{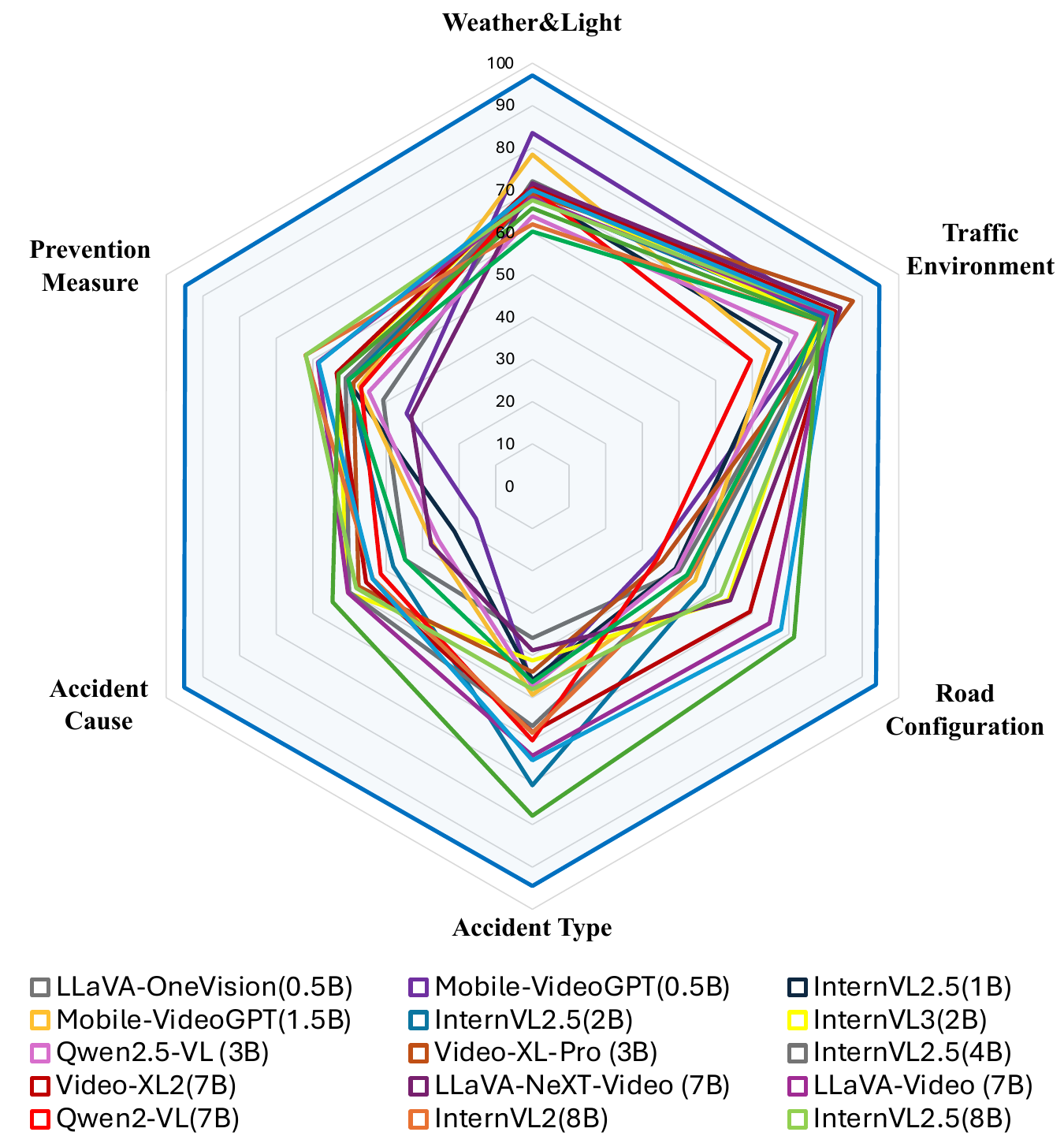}
\centering
\vspace{-1em}
\caption{Summary of MLLMs performance on VRU-Accident for the VQA task along with the Human Expert scores.}
\label{fig:pychart}
\end{figure}

\begin{table*}[htbp]
    \centering
    \caption{Quantitative comparisons on the dense caption task. We report ROUGE precision (P), recall (R), and F-measure (F), where P and R indicate the overlap of 4-grams with candidate and reference summaries, respectively, and F is their harmonic mean. Higher values indicate better performance. \textbf{Black}, \textbf{\textcolor{blue}{blue}}, and \textbf{\textcolor{red}{red}} colors indicate the best, second-best, and worst performance, respectively. \textbf{\underline{Underline}} denotes the best performance among open-source models.}
     \resizebox{\textwidth}{!}{
    \begin{tabular}{l|c|c|c|ccc |ccc | ccc}
    \hline
\multirow{2}{*} {Model (Param.)} & \multirow{2}{*}{SPICE $\uparrow$} & \multirow{2}{*}{METEOR $\uparrow$}& \multirow{2}{*} {COMET $\uparrow$} &  \multicolumn{3}{c|}{ROUGE-1 $\uparrow$ } & \multicolumn{3}{c|}{ROUGE-2 $\uparrow$} & \multicolumn{3}{c}{ROUGE-L $\uparrow$} \\
 &  &  & &  P & R & F & P & R & F & P & R & F\\
\hline\hline
LLaVA-OneVision(0.5B) \cite{LLaVA_OneVision} 
& 0.126
& 	0.224
& 0.647
&  0.388
&  0.384
&  0.380
& 0.092
& 0.091
& 0.090
&0.228
& 0.229
& 0.225
\\
 InternVL2.5(1B) \cite{InternVL25} 
& 0.132
 & 	0.236
 & 0.676
 &  0.409
&  0.403
 &  0.400
 & 0.089
 & 0.088
 & 0.087
& 0.212
 & 0.209
 & 0.207
\\
 Mobile-VideoGPT(1.5B) \cite{Mobile_VideoGPT} 
& 0.128
 &  0.25
 & 0.665
 &  \textbf{\textcolor{red}{0.363}}
 &  0.420
 & 0.386
 & \textbf{\textcolor{red}{0.069}}
 & 0.080
 & 0.073
 & \textbf{\textcolor{red}{0.180}}
 & 0.209
  & \textbf{\textcolor{red}{0.192}}
 \\
 InternVL2.5(2B) \cite{InternVL25} 
& 0.139
 & 0.249
 & 0.682
 & 0.404
& 0.428
 & 0.411
 & 0.094
 & 0.099
 & 0.095
& 0.207
 & 0.220
 & 0.211
\\
InternVL3(2B) \cite{InternVL3} 
& \textbf{\textcolor{red}{0.100}}
 & 	\textbf{\textcolor{red}{0.188}}
 & \textbf{\textcolor{red}{0.584}}
 &  0.429
 &  \textbf{\textcolor{red}{0.309}}
 &  \textbf{\textcolor{red}{0.345}}
 & 0.098
 & \textbf{\textcolor{red}{0.070}}
 & \textbf{\textcolor{red}{0.078}}
 & 0.249
 & \textbf{\textcolor{red}{0.177}}
 & 0.199
 \\
Qwen2.5-VL(3B) \cite{qwen25_VL} 
& 0.149
 &  0.261
 & 0.682
 &  0.396
 &  0.446
 &  0.416
 &  0.106
 &  0.120
 &  0.111
 &  0.207
 &  0.234
 &  0.218
 \\
Video-XL-Pro(3B) \cite{Video_XL_Pro} 
&  0.141
 &  0.265
 & 0.695
 &  0.396
 &  0.450
 &  0.418
 &  0.098
 &  0.111
 &  0.103
 &  0.206
 &  0.236
 &  0.218
 \\
 InternVL2.5(4B) \cite{InternVL25} 
&  0.146
 &  0.260
 & 0.696
 &  0.408
 &  0.449
 &   0.423
 &  0.098
 &  0.108
 &  0.102
 &  0.207
 &  0.228
 &  0.215
 \\
 Video-XL2(7B) \cite{Video_XL2} 
&  0.148
 &  0.233
 & 0.693
 &  \textbf{\textcolor{blue}{\underline{0.491}}}
 &  0.378
 &   0.421
 &  0.121
 &  0.092
 &  0.103
 &  \textbf{\textcolor{blue}{\underline{0.266}}}
 &  0.203
 &  0.227
 \\
 LLaVA-NeXT-Video(7B) \cite{LLaVA_NeXT_Video} 
&  0.155
 &  0.259
 & 0.708
 &  0.443
 &  0.438
 &   0.437
 &  0.124
 &  0.123
 &  0.123
 &  0.248
 &  0.246
 &  0.245
 \\
 LLaVA-Video(7B) \cite{LLaVA_Video} 
&  0.157
 &  \textbf{\textcolor{blue}{0.270}}
 & 0.686
 &  0.422
 &  0.460
 &   0.436
 &  0.113
 &  0.124
 &  0.117
 &  0.219
 &  0.240
 &  0.227
 \\
 Qwen2-VL(7B) \cite{qwen2_VL} 
&  \textbf{\textcolor{blue}{\underline{0.170}}}
 &  \textbf{\underline{0.285}}
 & \textbf{\textcolor{blue}{\underline{0.721}}}
 &  0.430
 &  \textbf{\underline{0.494}}
 &   \textbf{\textcolor{blue}{\underline{0.456}}}
 &  \textbf{\textcolor{blue}{\underline{0.126}}}
 &  \textbf{\textcolor{blue}{\underline{0.145}}}
 &  \textbf{\textcolor{blue}{\underline{0.134}}}
 &  0.233
 &  \textbf{\underline{0.270}}
 &  \textbf{\textcolor{blue}{\underline{0.248}}}
 \\
 InternVL2(8B) \cite{InternVL2} 
&  0.139
 &  0.249
 & 0.694
 &  0.398
 &  0.431
 &   0.410
 &  0.091
 &  0.098
 &  0.093
 &  0.201
 &  0.220
 &  0.208
 \\
 InternVL2.5(8B) \cite{InternVL25} 
&  0.152
 &  0.259
 & 0.698
 &  0.420
 &  0.442
 &   0.426
 &  0.102
 &  0.108
 &  0.104
 &  0.216
 &  0.228
 &  0.219
 \\
 InternVL3(8B) \cite{InternVL3} 
&  0.159
 &  0.267
 & 0.694
 &  0.434
 &  0.454
 &   0.437
 &  0.113
 &  0.118
 &  0.114
 &  0.225
 &  0.234
 &  0.225
 \\
 \hline
 \rowcolor{gray!15}
Gemini 1.5-flash \cite{gemini} 
&  \textbf{0.194}
 &  \textbf{0.285}
 & \textbf{0.741}
 &  \textbf{0.492}
 &  \textbf{\textcolor{blue}{0.479}}
 &   \textbf{0.481}
 &  \textbf{0.152}
 &  \textbf{0.147}
 &  \textbf{0.148}
 &  \textbf{0.272}
 &  \textbf{\textcolor{blue}{0.266}}
 &  \textbf{0.266}
 \\
\hline
    \end{tabular}
    
        \label{tab:performance}}
\end{table*}

\begin{figure*}[!h]
    \centerline{\includegraphics[width=\textwidth]{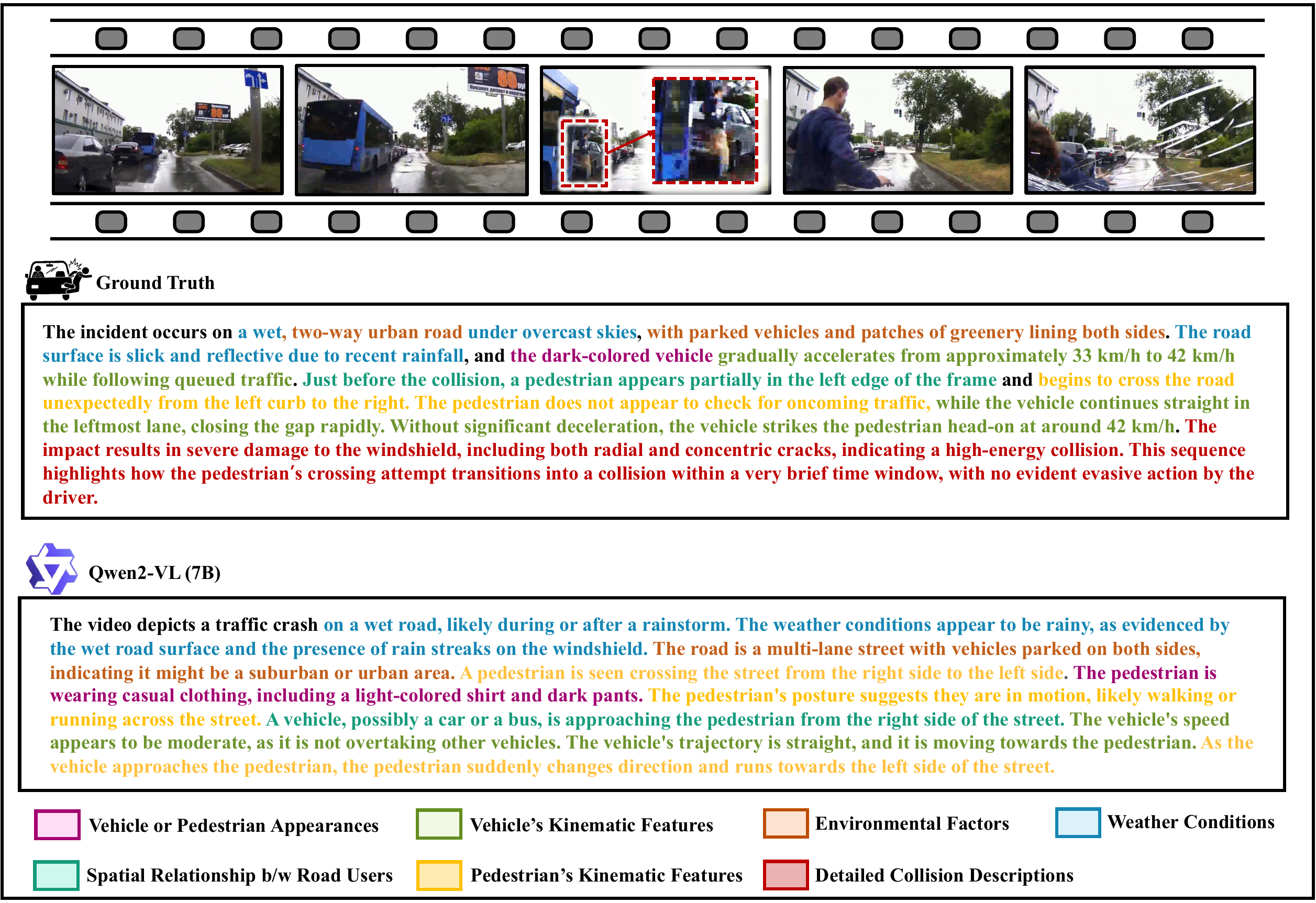}}
    
    \caption{Accident description from both Ground Truth in VRU-Accident and Qwen2-VL(7B) which achieved the best performance in the dense captioning task. Please refer to Figure~\ref{fig:Example_Visualization} in Supplementary Section~\ref{sec:example} for descriptions generated from all MLLMs.}
    \label{fig:Dense_Captioning_Visualization} 
\end{figure*}

\subsection{Result Analysis}

\textbf{VQA Task.} 
Table~\ref{tab:VQA} presents the performance comparison of 17 state-of-the-art MLLMs, including 15 open-source and 2 closed-source models, on the VQA task of the VRU-Accident benchmark. Overall, Gemini~\cite{gemini} achieves the best average performance (66.9\%), followed by LLaVA-Video~\cite{LLaVA_Video} and InternVL3(8B)~\cite{InternVL3}, which show strong performance across most categories. In contrast, smaller models such as Qwen2.5-VL(3B)~\cite{qwen25_VL} and InternVL2.5(1B)~\cite{InternVL25} struggle particularly on causal and preventive reasoning. We observe that MLLMs generally perform better on visually grounded attributes (e.g., Weather \& Light or Traffic Environment), while achieving lower accuracy on high-level reasoning tasks such as Accident Cause and Prevention Measure. Additionally, models often struggle with categories like `Road Configuration' and `Accident Type', but for distinct reasons. In the case of `Road Configuration', poor performance is largely due to the models’ limited ability to understand complex and dynamic properties of the surrounding environment, such as curves, intersections, or multi-lane structures. For `Accident Type', the difficulty stems from the models’ inability to reliably distinguish between different types of accident agents. For instance, models frequently misclassify options when both a pedestrian and a cyclist are mentioned, or when they must differentiate between ego-vehicle and surrounding vehicles, indicating limitations in fine-grained object-level reasoning within accident contexts. Notably, performance in the `Accident Cause' category is consistently poor across most models, with several falling below 30\%, indicating that current MLLMs lack robust causal reasoning capabilities in safety-critical scenarios. Compared to human experts, who achieve over 94\% accuracy in all categories, the gap highlights significant room for improvement in accident-centric understanding. These findings emphasize the importance of including causal and preventive questions in benchmarks for evaluating deeper vision-language reasoning, particularly in scenarios involving VRUs. The proposed VRU-Accident benchmark thus provides a challenging and diagnostic testbed for probing the limitations and capabilities of modern MLLMs in accident-related understanding.

To provide a holistic overview of model performance of VQA task, we visualize the results in Figure~\ref{fig:pychart}. This summary highlights the overall limitations of current MLLMs when applied to complex, safety-critical accident scenarios. Despite strong performance on certain visual attributes, most models struggle with causal reasoning and fine-grained collision understanding. These findings underscore the challenging nature of this benchmark and indicate substantial room for improvement in developing more robust and context-aware multimodal models.

\textbf{Dense Captioning Task.}
Table \ref{tab:performance} benchmarks 16 SOTA VLMs on SPICE, METEOR, COMET, and ROUGE. GPT-4 is excluded from this comparison, as it was used to generate the initial reference captions for the videos in the VRU-Accident benchmark. Gemini 1.5-flash leads the leaderboard, recording 0.194 SPICE, 0.285 METEOR, and 0.741 COMET, together with the highest ROUGE-F values of 0.481 (R-1), 0.148 (R-2) and 0.266 (R-L). Compared with the best open-source VLM, Qwen2-VL (7B), Gemini has absolute gains of +0.024 SPICE, +0.020 COMET, +0.025 ROUGE-1 F, +0.014 ROUGE-2 F and +0.018 ROUGE-L F, while matching Qwen2-VL on METEOR. These margins show that Gemini extracts finer scene details and maintains better temporal coverage in complex accident videos. Among open-source VLMs, Qwen2-VL (7B) is the clear front-runner. It couples the highest open-source scores in METEOR (0.285) and COMET (0.721) with strong balance between precision and recall, delivering 0.456/0.134/0.248 ROUGE-F. When precision alone is considered, Video-XL2 (7B) edges ahead with 0.491 (R-1 P) and 0.266 (R-L P), yet its low recall (0.378 R-1) indicates it produces short captions that omit salient context. In contrast, InternVL3 (2B) delivers the weakest semantic quality (0.100 SPICE, 0.584 COMET) and the lowest ROUGE-F (0.345 R-1), suggesting fragmented, short descriptions. Mobile-VideoGPT (1.5B) also under-performs, posting the smallest ROUGE-1 precision (0.363) and ROUGE-L precision (0.180), consistent with captions that miss core entities and actions. Overall, Gemini 1.5-flash retains the lead across both the VQA and dense-caption tasks on VRU-Accident. For open-source workflows, Qwen2-VL (7B) is the best choice for producing rich, spatiotemporal dynamics aware accident descriptions. \emph{Overall}, although Gemini 1.5-flash and Qwen2-VL (7B) achieve high COMET score, their low SPICE, METEOR, and ROUGE-L scores reveal clear gaps in causal reasoning, spatial relation understanding and temporal coherence.

We present a qualitative comparison between the ground truth caption in VRU-Accident and the caption generated by Qwen2-VL (7B), the best-performing open-source model in the dense captioning task, as shown in Figure~\ref{fig:Dense_Captioning_Visualization}. Qwen2-VL (7B) captures key contextual elements, such as weather conditions, environmental layout, pedestrian appearances, and vehicle trajectories, which contributes to its strong overall performance across the evaluation metrics. However, despite these strengths, the model fails to describe critical details about the collision itself, an essential component in safety-critical scenarios, and even hallucinates the pedestrian’s motion, misrepresenting their crossing direction. This highlights the inherent challenge of the VRU-Accident benchmark, which requires not only general scene understanding but also fine-grained reasoning about spatiotemporal dynamics in accident scenarios.
\section{Conclusion}
\label{sec:conclusion}

This work presents VRU-Accident, the first large-scale vision-language benchmark specifically designed to evaluate multimodal large language models (MLLMs) in safety-critical scenarios involving VRUs. Our benchmark comprises 1K real-world traffic accident videos annotated with over 6K question-answer pairs and 1K dense descriptions, enabling fine-grained assessments of both causal reasoning and narrative generation in accident contexts. Through extensive evaluations of 17 state-of-the-art MLLMs, we found that while models perform well on visually grounded attributes, they struggle with high-level reasoning tasks such as accident causes, prevention measures, road configurations, and accident type identification. These limitations highlight the current models’ difficulty in interpreting dynamic and complex environmental cues, as well as distinguishing between fine-grained object roles in collision events. By offering both diagnostic and generative evaluation settings, VRU-Accident provides a unified and challenging testbed to advance the capabilities of future multimodal models for the safe navigation of autonomous driving systems.


{
    \small
    \bibliographystyle{ieeenat_fullname}
    \bibliography{main}

\begin{thebibliography}{57}
\providecommand{\natexlab}[1]{#1}
\providecommand{\url}[1]{\texttt{#1}}
\expandafter\ifx\csname urlstyle\endcsname\relax
  \providecommand{\doi}[1]{doi: #1}\else
  \providecommand{\doi}{doi: \begingroup \urlstyle{rm}\Url}\fi

\bibitem[Abdelrahman et~al.(2025{\natexlab{a}})Abdelrahman, Abdel-Aty, and Tran]{abdelrahman2025vru}
Ahmed~S Abdelrahman, Mohamed Abdel-Aty, and Quoc~Dai Tran.
\newblock Vru-cipi: Crossing intention prediction at intersections for improving vulnerable road users safety.
\newblock In \emph{Proceedings of the Computer Vision and Pattern Recognition Conference}, pages 5623--5632, 2025{\natexlab{a}}.

\bibitem[Abdelrahman et~al.(2025{\natexlab{b}})Abdelrahman, Abdel-Aty, and Wang]{abdelrahman2025video}
Ahmed~S Abdelrahman, Mohamed Abdel-Aty, and Dongdong Wang.
\newblock Video-to-text pedestrian monitoring (vtpm): Leveraging large language models for privacy-preserve pedestrian activity monitoring at intersections.
\newblock In \emph{Proceedings of the Winter Conference on Applications of Computer Vision}, pages 366--375, 2025{\natexlab{b}}.

\bibitem[Abdelrahman et~al.(2025{\natexlab{c}})Abdelrahman, Abdel-Aty, Yang, and Faden]{abdelrahman2025advanced}
Ahmed~S Abdelrahman, Mohamed Abdel-Aty, Samgyu Yang, and Abdulrahman Faden.
\newblock Advanced crash causation analysis for freeway safety: A large language model approach to identifying key contributing factors.
\newblock \emph{arXiv preprint arXiv:2505.09949}, 2025{\natexlab{c}}.

\bibitem[Abdelrahman et~al.(2025{\natexlab{d}})Abdelrahman, Islam, and Abdel-Aty]{abdelrahman2025vrucrosssafe}
Ahmed~S Abdelrahman, Zubayer Islam, and Mohamed Abdel-Aty.
\newblock Vrucrosssafe for crossing intention prediction of vulnerable road users for improving safe crossing at intersections.
\newblock \emph{npj Sustainable Mobility and Transport}, 2\penalty0 (1):\penalty0 201--15, 2025{\natexlab{d}}.

\bibitem[Ahsan et~al.(2025)Ahsan, Abdel-Aty, and Abdelrahman]{ahsan2025evaluating}
Md~Jamil Ahsan, Mohamed Abdel-Aty, and Ahmed~S Abdelrahman.
\newblock Evaluating the safety impact of mid-block pedestrian signals (mps).
\newblock \emph{Accident Analysis \& Prevention}, 210:\penalty0 107847, 2025.

\bibitem[Anderson et~al.(2016)Anderson, Fernando, Johnson, and Gould]{SPICE}
Peter Anderson, Basura Fernando, Mark Johnson, and Stephen Gould.
\newblock Spice: Semantic propositional image caption evaluation.
\newblock In \emph{Computer Vision--ECCV 2016: 14th European Conference, Amsterdam, The Netherlands, October 11-14, 2016, Proceedings, Part V 14}, pages 382--398. Springer, 2016.

\bibitem[Chen et~al.(2024)Chen, Wang, Tian, Ye, Gao, Cui, Tong, Hu, Luo, Ma, et~al.]{InternVL2}
Zhe Chen, Weiyun Wang, Hao Tian, Shenglong Ye, Zhangwei Gao, Erfei Cui, Wenwen Tong, Kongzhi Hu, Jiapeng Luo, Zheng Ma, et~al.
\newblock How far are we to gpt-4v? closing the gap to commercial multimodal models with open-source suites.
\newblock \emph{arXiv preprint arXiv:2404.16821}, 2024.

\bibitem[Chen et~al.(2025)Chen, Wang, Cao, Liu, Gao, Cui, Zhu, Ye, Tian, Liu, Gu, Wang, Li, Ren, Chen, Luo, Wang, Jiang, Wang, He, Shi, Zhang, Lv, Wang, Shao, Chu, Tu, He, Wu, Deng, Ge, Chen, Zhang, Wang, Dou, Lu, Zhu, Lu, Lin, Qiao, Dai, and Wang]{InternVL25}
Zhe Chen, Weiyun Wang, Yue Cao, Yangzhou Liu, Zhangwei Gao, Erfei Cui, Jinguo Zhu, Shenglong Ye, Hao Tian, Zhaoyang Liu, Lixin Gu, Xuehui Wang, Qingyun Li, Yimin Ren, Zixuan Chen, Jiapeng Luo, Jiahao Wang, Tan Jiang, Bo Wang, Conghui He, Botian Shi, Xingcheng Zhang, Han Lv, Yi Wang, Wenqi Shao, Pei Chu, Zhongying Tu, Tong He, Zhiyong Wu, Huipeng Deng, Jiaye Ge, Kai Chen, Kaipeng Zhang, Limin Wang, Min Dou, Lewei Lu, Xizhou Zhu, Tong Lu, Dahua Lin, Yu Qiao, Jifeng Dai, and Wenhai Wang.
\newblock Expanding performance boundaries of open-source multimodal models with model, data, and test-time scaling, 2025.

\bibitem[Datta et~al.(2022)Datta, Joshi, and Gupta]{datta2022analysis}
Goutam Datta, Nisheeth Joshi, and Kusum Gupta.
\newblock Analysis of automatic evaluation metric on low-resourced language: Bertscore vs bleu score.
\newblock In \emph{International Conference on Speech and Computer}, pages 155--162. Springer, 2022.

\bibitem[Fang et~al.(2019)Fang, Yan, Qiao, Xue, Wang, and Li]{DADA_2000}
Jianwu Fang, Dingxin Yan, Jiahuan Qiao, Jianru Xue, He Wang, and Sen Li.
\newblock Dada-2000: Can driving accident be predicted by driver attention? analyzed by a benchmark, 2019.

\bibitem[Fang et~al.(2023)Fang, Li, Yang, Zheng, Xue, and Chua]{CAP_DATA}
Jianwu Fang, Lei-Lei Li, Kuan Yang, Zhedong Zheng, Jianru Xue, and Tat-Seng Chua.
\newblock Cognitive accident prediction in driving scenes: A multimodality benchmark, 2023.

\bibitem[Fang et~al.(2024)Fang, lei Li, Zhou, Xiao, Yu, Lv, Xue, and Chua]{MM_AU}
Jianwu Fang, Lei lei Li, Junfei Zhou, Junbin Xiao, Hongkai Yu, Chen Lv, Jianru Xue, and Tat-Seng Chua.
\newblock Abductive ego-view accident video understanding for safe driving perception, 2024.

\bibitem[(GHSA)(2020)]{GHSA2020}
Governors Highway Safety~Association (GHSA).
\newblock Pedestrian traffic fatalities by state: 2019 preliminary data, 2020.

\bibitem[Gopalkrishnan et~al.(2024)Gopalkrishnan, Greer, and Trivedi]{MLLM1}
Akshay Gopalkrishnan, Ross Greer, and Mohan Trivedi.
\newblock Multi-frame, lightweight \& efficient vision-language models for question answering in autonomous driving, 2024.

\bibitem[Ham et~al.(2023)Ham, Kim, Jung, and Moon]{ham2023cipf}
Je-Seok Ham, Dae~Hoe Kim, NamKyo Jung, and Jinyoung Moon.
\newblock Cipf: Crossing intention prediction network based on feature fusion modules for improving pedestrian safety.
\newblock In \emph{Proceedings of the IEEE/CVF Conference on Computer Vision and Pattern Recognition}, pages 3666--3675, 2023.

\bibitem[Karim et~al.(2024)Karim, Yin, and Qin]{ROL}
Muhammad~Monjurul Karim, Zhaozheng Yin, and Ruwen Qin.
\newblock An attention-guided multistream feature fusion network for early localization of risky traffic agents in driving videos.
\newblock \emph{IEEE Transactions on Intelligent Vehicles}, 9\penalty0 (1):\penalty0 1792--1803, 2024.

\bibitem[Kim et~al.(2018)Kim, Rohrbach, Darrell, Canny, and Akata]{TrafficQA11}
Jinkyu Kim, Anna Rohrbach, Trevor Darrell, John Canny, and Zeynep Akata.
\newblock Textual explanations for self-driving vehicles.
\newblock In \emph{Proceedings of the European conference on computer vision (ECCV)}, pages 563--578, 2018.

\bibitem[Kim et~al.(2025)Kim, Abdel-Aty, Choi, Islam, Wang, and Zhai]{VRU1}
Younggun Kim, Mohamed Abdel-Aty, Keechoo Choi, Zubayer Islam, Dongdong Wang, and Shaoyan Zhai.
\newblock Pedestrian crossing direction prediction at intersections for pedestrian safety.
\newblock \emph{IEEE Open Journal of Intelligent Transportation Systems}, 6:\penalty0 692--707, 2025.

\bibitem[Lavie and Agarwal(2007)]{METEOR}
Alon Lavie and Abhaya Agarwal.
\newblock Meteor: an automatic metric for mt evaluation with high levels of correlation with human judgments.
\newblock In \emph{Proceedings of the Second Workshop on Statistical Machine Translation}, page 228–231, USA, 2007. Association for Computational Linguistics.

\bibitem[Li et~al.(2024{\natexlab{a}})Li, Zhang, Guo, Zhang, Li, Zhang, Zhang, Li, Liu, and Li]{LLaVA_OneVision}
Bo Li, Yuanhan Zhang, Dong Guo, Renrui Zhang, Feng Li, Hao Zhang, Kaichen Zhang, Yanwei Li, Ziwei Liu, and Chunyuan Li.
\newblock Llava-onevision: Easy visual task transfer, 2024{\natexlab{a}}.

\bibitem[Li et~al.(2024{\natexlab{b}})Li, Zhang, Zhang, Zhang, Li, Li, Ma, and Li]{LLaVA_NeXT_Video}
Feng Li, Renrui Zhang, Hao Zhang, Yuanhan Zhang, Bo Li, Wei Li, Zejun Ma, and Chunyuan Li.
\newblock Llava-next-interleave: Tackling multi-image, video, and 3d in large multimodal models, 2024{\natexlab{b}}.

\bibitem[Lin(2004)]{ROUGE}
Chin-Yew Lin.
\newblock {ROUGE}: A package for automatic evaluation of summaries.
\newblock In \emph{Text Summarization Branches Out}, pages 74--81, Barcelona, Spain, 2004. Association for Computational Linguistics.

\bibitem[Liu et~al.(2025{\natexlab{a}})Liu, Li, and Yin]{MLLM5}
Tianming Liu, Manzi Li, and Yafeng Yin.
\newblock Aligning llm with human travel choices: a persona-based embedding learning approach, 2025{\natexlab{a}}.

\bibitem[Liu et~al.(2025{\natexlab{b}})Liu, Yang, and Yin]{MLLM3}
Tianming Liu, Jirong Yang, and Yafeng Yin.
\newblock Toward llm-agent-based modeling of transportation systems: A conceptual framework.
\newblock \emph{Artificial Intelligence for Transportation}, 1:\penalty0 100001, 2025{\natexlab{b}}.

\bibitem[Liu et~al.(2025{\natexlab{c}})Liu, Shu, Liu, Li, Tian, and Zhao]{Video_XL_Pro}
Xiangrui Liu, Yan Shu, Zheng Liu, Ao Li, Yang Tian, and Bo Zhao.
\newblock Video-xl-pro: Reconstructive token compression for extremely long video understanding, 2025{\natexlab{c}}.

\bibitem[Luo and Wang(2023)]{CTAD}
Haohan Luo and Feng Wang.
\newblock A simulation-based framework for urban traffic accident detection.
\newblock In \emph{ICASSP 2023 - 2023 IEEE International Conference on Acoustics, Speech and Signal Processing (ICASSP)}, pages 1--5, 2023.

\bibitem[Ma et~al.(2024)Ma, Cao, Sun, Pavone, and Xiao]{MLLM2}
Yingzi Ma, Yulong Cao, Jiachen Sun, Marco Pavone, and Chaowei Xiao.
\newblock Dolphins: Multimodal language model for driving.
\newblock In \emph{European Conference on Computer Vision}, pages 403--420. Springer, 2024.

\bibitem[Malla et~al.(2023)Malla, Choi, Dwivedi, Choi, and Li]{TrafficQA7}
Srikanth Malla, Chiho Choi, Isht Dwivedi, Joon~Hee Choi, and Jiachen Li.
\newblock Drama: Joint risk localization and captioning in driving.
\newblock In \emph{Proceedings of the IEEE/CVF winter conference on applications of computer vision}, pages 1043--1052, 2023.

\bibitem[Marcu et~al.(2023)Marcu, Chen, Hünermann, Karnsund, Hanotte, Chidananda, Nair, Badrinarayanan, Kendall, Shotton, and Sinavski]{TrafficQA2}
Ana-Maria Marcu, Long Chen, Jan Hünermann, Alice Karnsund, Benoit Hanotte, Prajwal Chidananda, Saurabh Nair, Vijay Badrinarayanan, Alex Kendall, Jamie Shotton, and Oleg Sinavski.
\newblock Lingoqa: Visual question answering for autonomous driving.
\newblock \emph{arXiv preprint arXiv:2312.14115}, 2023.

\bibitem[OpenAI(2024)]{GPT}
OpenAI.
\newblock Gpt-4 technical report, 2024.

\bibitem[Parikh et~al.(2024)Parikh, Saluja, Jawahar, and Sarvadevabhatla]{TrafficQA8}
Chirag Parikh, Rohit Saluja, CV Jawahar, and Ravi~Kiran Sarvadevabhatla.
\newblock Idd-x: A multi-view dataset for ego-relative important object localization and explanation in dense and unstructured traffic.
\newblock In \emph{2024 IEEE International Conference on Robotics and Automation (ICRA)}, pages 14815--14821. IEEE, 2024.

\bibitem[Qasemi et~al.(2023)Qasemi, Francis, and Oltramari]{TrafficQA3}
Ehsan Qasemi, Jonathan~M Francis, and Alessandro Oltramari.
\newblock Traffic-domain video question answering with automatic captioning.
\newblock \emph{arXiv preprint arXiv:2307.09636}, 2023.

\bibitem[Qian et~al.(2023)Qian, Chen, Zhuo, Jiao, and Jiang]{TrafficQA1}
Tianwen Qian, Jingjing Chen, Linhai Zhuo, Yang Jiao, and Yu-Gang Jiang.
\newblock Nuscenes-qa: A multi-modal visual question answering benchmark for autonomous driving scenario.
\newblock \emph{arXiv preprint arXiv:2305.14836}, 2023.

\bibitem[Rei et~al.(2020)Rei, Stewart, Farinha, and Lavie]{COMET}
Ricardo Rei, Craig Stewart, Ana~C Farinha, and Alon Lavie.
\newblock Comet: A neural framework for mt evaluation, 2020.

\bibitem[Sachdeva et~al.(2024)Sachdeva, Agarwal, Chundi, Roelofs, Li, Kochenderfer, Choi, and Dariush]{TrafficQA9}
Enna Sachdeva, Nakul Agarwal, Suhas Chundi, Sean Roelofs, Jiachen Li, Mykel Kochenderfer, Chiho Choi, and Behzad Dariush.
\newblock Rank2tell: A multimodal driving dataset for joint importance ranking and reasoning.
\newblock In \emph{Proceedings of the IEEE/CVF winter conference on applications of computer vision}, pages 7513--7522, 2024.

\bibitem[Shaker et~al.(2025)Shaker, Maaz, Rezatofighi, Khan, and Khan]{Mobile_VideoGPT}
Abdelrahman Shaker, Muhammad Maaz, Hamid Rezatofighi, Salman Khan, and Fahad~Shahbaz Khan.
\newblock Mobile-videogpt: Fast and accurate video understanding language model.
\newblock \emph{arxiv}, 2025.

\bibitem[Shu et~al.(2024)Shu, Zhang, Liu, Qin, Zhou, Huang, and Zhao]{Video_XL2}
Yan Shu, Peitian Zhang, Zheng Liu, Minghao Qin, Junjie Zhou, Tiejun Huang, and Bo Zhao.
\newblock Video-xl: Extra-long vision language model for hour-scale video understanding.
\newblock \emph{arXiv preprint arXiv:2409.14485}, 2024.

\bibitem[Sima et~al.(2023)Sima, Renz, Chitta, Chen, Zhang, Xie, Luo, Geiger, and Li]{TrafficQA4}
Chonghao Sima, Katrin Renz, Kashyap Chitta, Li Chen, Hanxue Zhang, Chengen Xie, Ping Luo, Andreas Geiger, and Hongyang Li.
\newblock Drivelm: Driving with graph visual question answering.
\newblock \emph{arXiv preprint arXiv:2312.14150}, 2023.

\bibitem[Team(2025{\natexlab{a}})]{gemini}
Gemini Team.
\newblock Gemini: A family of highly capable multimodal models, 2025{\natexlab{a}}.

\bibitem[Team(2025{\natexlab{b}})]{qwen25_VL}
Qwen Team.
\newblock Qwen2.5-vl, 2025{\natexlab{b}}.

\bibitem[Wang et~al.(2022)Wang, Chen, Zhang, Wang, Yu, and Cheng]{Wang1}
Chenzhu Wang, Fei Chen, Yunlong Zhang, Shuyi Wang, Bin Yu, and Jianchuan Cheng.
\newblock Temporal stability of factors affecting injury severity in rear-end and non-rear-end crashes: A random parameter approach with heterogeneity in means and variances.
\newblock \emph{Analytic Methods in Accident Research}, 35:\penalty0 100219, 2022.

\bibitem[Wang et~al.(2024{\natexlab{a}})Wang, Abdel-Aty, and Han]{Lei2}
Chenzhu Wang, Mohamed Abdel-Aty, and Lei Han.
\newblock Effects of speed difference on injury severity of freeway rear-end crashes: Insights from correlated joint random parameters bivariate probit models and temporal instability.
\newblock \emph{Analytic Methods in Accident Research}, 42:\penalty0 100320, 2024{\natexlab{a}}.

\bibitem[Wang et~al.(2024{\natexlab{b}})Wang, Abdel-Aty, Han, and Easa]{Lei3}
Chenzhu Wang, Mohamed Abdel-Aty, Lei Han, and Said~M. Easa.
\newblock Analyzing speed-difference impact on freeway joint injury severities of leading-following vehicles using statistical and data-driven models.
\newblock \emph{Accident Analysis \& Prevention}, 206:\penalty0 107695, 2024{\natexlab{b}}.

\bibitem[Wang et~al.(2025)Wang, Abdel-Aty, and Han]{Lei1}
Chenzhu Wang, Mohamed Abdel-Aty, and Lei Han.
\newblock Tunnel crash severity and congestion duration joint evaluation based on cross-stitch networks.
\newblock \emph{Accident Analysis \& Prevention}, 213:\penalty0 107942, 2025.

\bibitem[Wang et~al.(2024{\natexlab{c}})Wang, Bai, Tan, Wang, Fan, Bai, Chen, Liu, Wang, Ge, Fan, Dang, Du, Ren, Men, Liu, Zhou, Zhou, and Lin]{qwen2_VL}
Peng Wang, Shuai Bai, Sinan Tan, Shijie Wang, Zhihao Fan, Jinze Bai, Keqin Chen, Xuejing Liu, Jialin Wang, Wenbin Ge, Yang Fan, Kai Dang, Mengfei Du, Xuancheng Ren, Rui Men, Dayiheng Liu, Chang Zhou, Jingren Zhou, and Junyang Lin.
\newblock Qwen2-vl: Enhancing vision-language model's perception of the world at any resolution.
\newblock \emph{arXiv preprint arXiv:2409.12191}, 2024{\natexlab{c}}.

\bibitem[Wang et~al.(2023)Wang, Kim, Ji, Xie, Ge, Chen, Li, and Luo]{DeepAccident}
Tianqi Wang, Sukmin Kim, Wenxuan Ji, Enze Xie, Chongjian Ge, Junsong Chen, Zhenguo Li, and Ping Luo.
\newblock Deepaccident: A motion and accident prediction benchmark for v2x autonomous driving, 2023.

\bibitem[Xu et~al.(2021)Xu, Huang, and Liu]{SUTD_TrafficQA}
Li Xu, He Huang, and Jun Liu.
\newblock Sutd-trafficqa: A question answering benchmark and an efficient network for video reasoning over traffic events, 2021.

\bibitem[Xu et~al.(2020)Xu, Yang, Gong, Lin, Wu, Li, and Vasconcelos]{TrafficQA10}
Yiran Xu, Xiaoyin Yang, Lihang Gong, Hsuan-Chu Lin, Tz-Ying Wu, Yunsheng Li, and Nuno Vasconcelos.
\newblock Explainable object-induced action decision for autonomous vehicles.
\newblock In \emph{Proceedings of the IEEE/CVF Conference on Computer Vision and Pattern Recognition}, pages 9523--9532, 2020.

\bibitem[Yao et~al.(2019)Yao, Xu, Wang, Crandall, and Atkins]{A3D}
Yu Yao, Mingze Xu, Yuchen Wang, David~J. Crandall, and Ella~M. Atkins.
\newblock Unsupervised traffic accident detection in first-person videos.
\newblock In \emph{2019 IEEE/RSJ International Conference on Intelligent Robots and Systems (IROS)}, pages 273--280, 2019.

\bibitem[Yao et~al.(2023)Yao, Wang, Xu, Pu, Wang, Atkins, and Crandall]{DoTA}
Yu Yao, Xizi Wang, Mingze Xu, Zelin Pu, Yuchen Wang, Ella Atkins, and David~J. Crandall.
\newblock Dota: Unsupervised detection of traffic anomaly in driving videos.
\newblock \emph{IEEE Transactions on Pattern Analysis and Machine Intelligence}, 45\penalty0 (1):\penalty0 444--459, 2023.

\bibitem[Yin and Zhang(2024)]{Similarity}
Chen Yin and Zixuan Zhang.
\newblock A study of sentence similarity based on the all-minilm-l6-v2 model with “same semantics, different structure” after fine tuning.
\newblock In \emph{Proceedings of the 2024 2nd International Conference on Image, Algorithms and Artificial Intelligence (ICIAAI 2024)}, pages 677--684. Atlantis Press, 2024.

\bibitem[You et~al.(2023)You, Zhang, Gan, Du, Zhang, Wang, Cao, Chang, and Yang]{you2023ferret}
Haoxuan You, Haotian Zhang, Zhe Gan, Xianzhi Du, Bowen Zhang, Zirui Wang, Liangliang Cao, Shih-Fu Chang, and Yinfei Yang.
\newblock Ferret: Refer and ground anything anywhere at any granularity.
\newblock \emph{arXiv preprint arXiv:2310.07704}, 2023.

\bibitem[You and Han(2020)]{CTA}
Tackgeun You and Bohyung Han.
\newblock {Traffic Accident Benchmark for Causality Recognition}.
\newblock In \emph{ECCV}, 2020.

\bibitem[Zhang et~al.(2024)Zhang, Wu, Li, Li, Ma, Liu, and Li]{LLaVA_Video}
Yuanhan Zhang, Jinming Wu, Wei Li, Bo Li, Zejun Ma, Ziwei Liu, and Chunyuan Li.
\newblock Video instruction tuning with synthetic data, 2024.

\bibitem[Zhou et~al.(2025)Zhou, Larintzakis, Guo, Zimmer, Liu, Cao, Zhang, Lakshminarasimhan, Strand, and Knoll]{TUMTraf-VideoQA}
Xingcheng Zhou, Konstantinos Larintzakis, Hao Guo, Walter Zimmer, Mingyu Liu, Hu Cao, Jiajie Zhang, Venkatnarayanan Lakshminarasimhan, Leah Strand, and Alois~C. Knoll.
\newblock {TUMTraffic}-{VideoQA}: A benchmark for unified spatio-temporal video understanding in traffic scenes.
\newblock \emph{IEEE/CVF Int. Conf. on Machine Learning (ICML)}, 2025.

\bibitem[Zhu et~al.(2025)Zhu, Wang, Chen, Liu, Ye, Gu, Tian, Duan, Su, Shao, Gao, Cui, Wang, Cao, Liu, Wei, Zhang, Wang, Xu, Li, Wang, Deng, Li, He, Jiang, Luo, Wang, He, Shi, Zhang, Shao, He, Xiong, Qu, Sun, Jiao, Lv, Wu, Zhang, Deng, Ge, Chen, Wang, Dou, Lu, Zhu, Lu, Lin, Qiao, Dai, and Wang]{InternVL3}
Jinguo Zhu, Weiyun Wang, Zhe Chen, Zhaoyang Liu, Shenglong Ye, Lixin Gu, Hao Tian, Yuchen Duan, Weijie Su, Jie Shao, Zhangwei Gao, Erfei Cui, Xuehui Wang, Yue Cao, Yangzhou Liu, Xingguang Wei, Hongjie Zhang, Haomin Wang, Weiye Xu, Hao Li, Jiahao Wang, Nianchen Deng, Songze Li, Yinan He, Tan Jiang, Jiapeng Luo, Yi Wang, Conghui He, Botian Shi, Xingcheng Zhang, Wenqi Shao, Junjun He, Yingtong Xiong, Wenwen Qu, Peng Sun, Penglong Jiao, Han Lv, Lijun Wu, Kaipeng Zhang, Huipeng Deng, Jiaye Ge, Kai Chen, Limin Wang, Min Dou, Lewei Lu, Xizhou Zhu, Tong Lu, Dahua Lin, Yu Qiao, Jifeng Dai, and Wenhai Wang.
\newblock Internvl3: Exploring advanced training and test-time recipes for open-source multimodal models, 2025.

\bibitem[Zimmer et~al.(2025)Zimmer, Greer, Zhou, Song, Cao, Lehmberg, Pavel, Alaaeldin~Ghita, Gopalkrishnan, Caesar, Trivedi, and Knoll]{TUMTraf-A}
Walter Zimmer, Ross Greer, Xingcheng Zhou, Rui Song, Hu Cao, Daniel Lehmberg, Marc Pavel, Ahmed Alaaeldin~Ghita, Akshay Gopalkrishnan, Holger Caesar, Mohan~M. Trivedi, and Alois~C. Knoll.
\newblock Towards vision zero: The accid3nd dataset.
\newblock page~10, 2025.

\end{thebibliography}
}

\newpage
\begin{figure}[t!]
\includegraphics[width=\columnwidth]{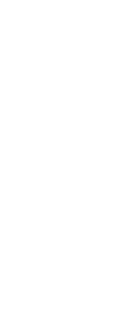}
\centering

\end{figure}
\appendix

\twocolumn[{%
  \maketitle
  \begin{center}
    {\Large \textbf{Supplementary: VRU-Accident: A Vision-Language Benchmark for Video Question Answering and Dense Captioning for Accident Scene Understanding}}
    \vspace{+1em}
    \end{center}
}]

\section{Supplementary Organization}
\label{sec:organization}

We organize the supplementary material as follows: 
\begin{itemize}
\item Section~\ref{sec:metrics}: VRU-Accident Details
\item Section~\ref{sec:example}: Qualitative Examples
\item Section~\ref{sec:prompt}: Prompts for VRU-Accident Curation
\item Section~\ref{sec:reproduction}: Reproduction of Experiment
\item Section~\ref{sec:acknowledgements}: Acknowledgements
\end{itemize}
 
\section{VRU-Accident Details}
\label{sec:metrics}

\subsection{Task-Specific Prompts}

We present the input prompts used for MLLMs for each of the two tasks in VRU-Accident: VQA and Dense Captioning. The prompts are designed to elicit informative and structured responses from the models when paired with video inputs.

\paragraph{Prompts for VQA task} 
For each question category in the VQA task, we provide a specific prompt.  The prompts used are:

\begin{itemize}
    \item \textbf{Weather and Light Condition:} What's the weather and lighting? Choose the correct option (A, B, C, or D) without any explanations.
    \item \textbf{Traffic Environment:} Where did the accident happen? Choose the correct option (A, B, C, or D) without any explanations.
    \item \textbf{Road Configuration:} What type of road is shown? Choose the correct option (A, B, C, or D) without any explanations.
    \item \textbf{Accident Type:} What kind of accident occurred? Choose the correct option (A, B, C, or D) without any explanations.
    \item \textbf{Accident Cause:} Why did the accident happen? Choose the correct option (A, B, C, or D) without any explanations.
    \item \textbf{Accident Prevention Measure:} How could this accident be prevented? Choose the correct option (A, B, C, or D) without any explanations.
\end{itemize}

\paragraph{Prompts for Dense Captioning task} 
To generate textual descriptions that match the level of detail and narrative style of the ground truth dense captions in VRU-Accident, we used the following prompt for all MLLMs:
\begin{itemize}
\item Provide a detailed description of this accident video. 
Use clear and complete sentences with appropriate traffic and accident-related terminology. 
Include descriptions of weather conditions, road type, and vehicle or pedestrian appearance (such as clothing and posture). 
Mention vehicle speed, trajectory, and movements, as well as any changes in the pedestrian's behavior. 
Focus on the dynamics of the collision, including vehicle approach, pedestrian movement, and final impact.
\end{itemize}

\subsection{Evaluation Metrics}
In this section, we supplement explanation of evaluation metrics used for both VQA and Dense Captioning tasks in the main manuscript.

\paragraph{VQA Evaluation Metrics} We evaluate the performance of models on the VQA task using standard classification accuracy per category. For each category $j$, we define the accuracy $\text{Acc}_j$ as $\text{Acc}_j = N^{\text{correct}}_j/N^{\text{total}}_j$ where $N^{\text{correct}}_j$ denotes the number of correctly predicted samples, and $N^{\text{total}}_j$ is the total number of samples for category $j$.
\textit{Note that}, unlike conventional classification tasks where a model selects from a fixed set of global class labels, each question $q_j$ in our benchmark is associated with a dynamically constructed candidate set $\hat{Y}_j$ that includes one correct answer and three counterfactual distractors. This makes the task more challenging, as models must perform fine-grained reasoning to distinguish the correct answer from contextually plausible but incorrect alternatives.

\paragraph{Dense Captioning Evaluation Metrics}
We select well-accepted and reliable metrics, including SPICE~\cite{SPICE}, METEOR~\cite{METEOR}, COMET~\cite{COMET}, and ROUGE scores~\cite{ROUGE}, to quantitatively evaluate the generated descriptions from the VLMs. 

First of all, the SPICE metric evaluates caption quality by comparing semantic content in the form of tuples extracted from scene graphs. Each caption is parsed into a set of tuples representing objects, attributes, and relations. The SPICE F1 score is computed as the harmonic mean of tuple-level precision and recall:

\begin{equation}
\text{SPICE}_{F1} = 2 \cdot \frac{\text{Precision} \cdot \text{Recall}}{\text{Precision} + \text{Recall}},
\end{equation}

\noindent where \text{Precision} and \text{Recall} are defined as:

\begin{align}
\text{Precision} &= \frac{|\mathcal{T}_{\text{gen}} \cap \mathcal{T}_{\text{ref}}|}{|\mathcal{T}_{\text{gen}}|}, \\
\text{Recall} &= \frac{|\mathcal{T}_{\text{gen}} \cap \mathcal{T}_{\text{ref}}|}{|\mathcal{T}_{\text{ref}}|},
\end{align}

\noindent with \( \mathcal{T}_{\text{gen}} \) and \( \mathcal{T}_{\text{ref}} \) denoting the sets of tuples from the generated and reference captions, respectively.

Secondly, METEOR captures both precision and recall of matching words, balancing linguistic precision and semantic recall. The METEOR score is calculated using the following formula:
\begin{equation}
   \text{METEOR} = \frac{\text{Precision} \times \text{Recall}}{\text{Precision} + \alpha \times \text{Recall} + (1 - \alpha)}
   \label{eq:meteor}
\end{equation}
Precision represents the proportion of words in the generated text that match the reference text, while Recall indicates the proportion of words in the reference text that are captured in the generated text. The parameter $\alpha$ functions as a weighting factor, balancing linguistic precision and semantic recall to provide an adaptive evaluation of the generated text's fidelity and coverage compared to the reference.

Thirdly, COMET is a neural-based metric that leverages pre-trained multilingual language models to predict human judgment scores for machine-generated text. Unlike surface-level metrics that rely solely on n-gram overlaps, COMET evaluates the semantic adequacy and fluency of the generated captions by comparing them against reference texts using contextual embeddings. Formally, COMET operates by encoding the source (\( S \)), reference (\( R \)), and hypothesis (\( H \)) using a pre-trained model and passing them through a regression head to produce a quality score:
\begin{equation}
    \text{COMET}(S, R, H) = f_{\theta}(\text{Enc}(S), \text{Enc}(R), \text{Enc}(H))
    \label{eq:comet}
\end{equation}
where \( \text{Enc}(\cdot) \) denotes the encoder output from the language model and \( f_{\theta} \) represents the learned regression layer that maps the embeddings to a scalar score. COMET provides a more human-aligned assessment of caption quality, especially in capturing nuanced semantic differences that traditional metrics may overlook.

Finally, the ROUGE score is a set of metrics used to evaluate the summarization quality, specifically through precision (P), recall (R), and harmonic mean (F) scores based on n-gram overlaps. We report ROUGE-1, ROUGE-2, and ROUGE-L scores, corresponding to unigram overlap, bigram overlap, and longest common subsequence (LCS), respectively. For each variant, we compute precision, recall, and F1 as follows:

\begin{align}
\text{Precision}_{\text{ROUGE-n}} &= \frac{|\text{n-gram}_{\text{gen}} \cap \text{n-gram}_{\text{ref}}|}{|\text{n-gram}_{\text{gen}}|}, \\
\text{Recall}_{\text{ROUGE-n}} &= \frac{|\text{n-gram}_{\text{gen}} \cap \text{n-gram}_{\text{ref}}|}{|\text{n-gram}_{\text{ref}}|}, \\
\text{F1}_{\text{ROUGE-n}} &= 2 \cdot \frac{\text{Precision} \cdot \text{Recall}}{\text{Precision} + \text{Recall}},
\end{align}

\noindent where \( n \in \{1, 2\} \) for ROUGE-1 and ROUGE-2, and \( \text{n-gram}_{\text{gen}} \), \( \text{n-gram}_{\text{ref}} \) denote the sets of n-grams in the generated and reference captions, respectively.

For ROUGE-L, which is based on the longest common subsequence (LCS), the scores are defined as:

\begin{align}
\text{Precision}_{\text{ROUGE-L}} &= \frac{\text{LCS}(\text{gen}, \text{ref})}{|\text{gen}|}, \\
\text{Recall}_{\text{ROUGE-L}} &= \frac{\text{LCS}(\text{gen}, \text{ref})}{|\text{ref}|}, \\
\text{F1}_{\text{ROUGE-L}} &= 2 \cdot \frac{\text{Precision} \cdot \text{Recall}}{\text{Precision} + \text{Recall}}.
\end{align}

\begin{figure*}[!t]
    \centerline{\includegraphics[width=\textwidth]{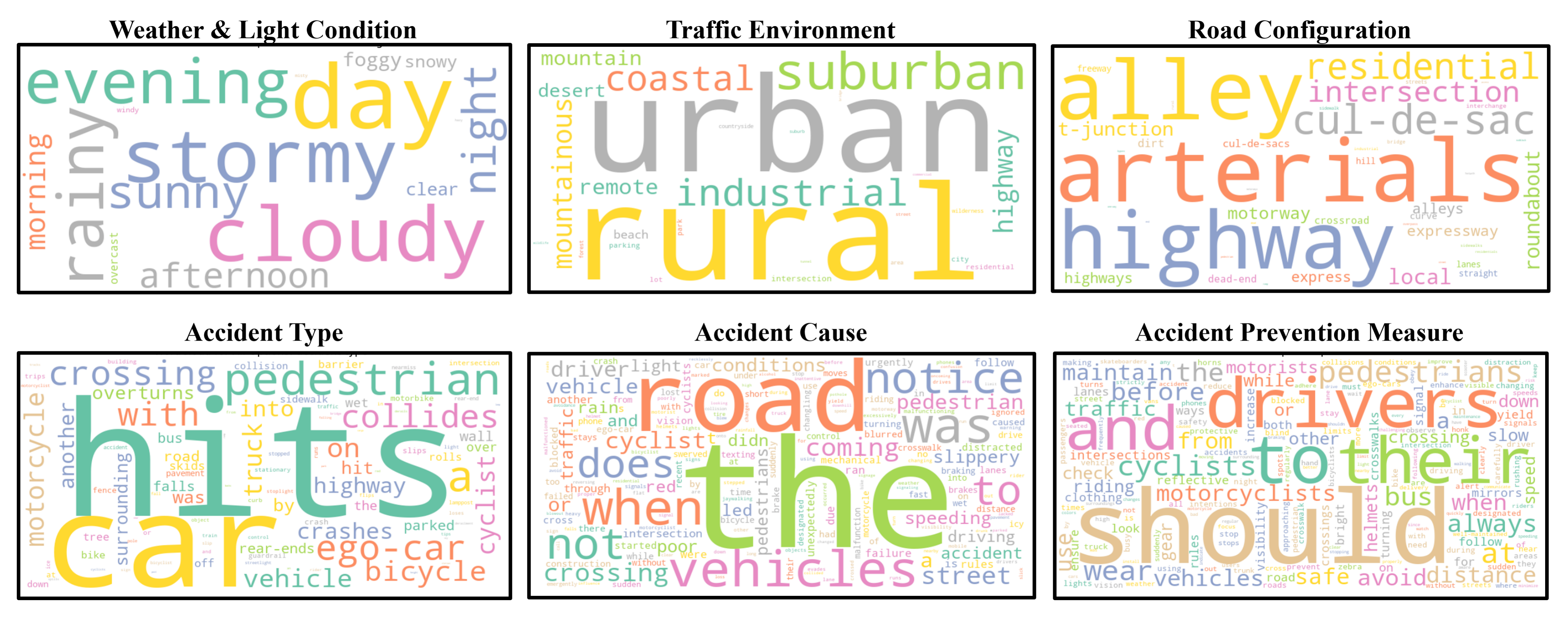}}
    \caption{Word frequency per each category in the VRU-Accident benchmark.}
    \label{fig:VQA_word_distribution}
\end{figure*}

\subsection{Annotation Detail}

Our VRU-Accident benchmark is designed to evaluate two core tasks: Video Question Answering (VQA) and Dense Captioning. The goal of the VQA task is to select the most appropriate answer from concise options—typically represented as short phrases a word, while the Dense Captioning task aims to generate detailed natural language descriptions that capture the full spatio-temporal context of a traffic accident. Given that real-world accident scenarios often involve complex and intertwined physical, environmental, and behavioral elements, it is inherently difficult to fully represent such events using concise options alone. Therefore, we provide annotation guidelines below that clarify how we determined the ground truth labels for each VQA category, while noting that richer, time-dependent contextual information is separately captured in the dense caption annotations.

\paragraph{Weather \& Light Condition}
This category is annotated based on how the weather and lighting conditions plausibly contributed to the accident. For instance, if visibility is generally clear and there is no evidence of severe shadows or rain, the condition is annotated as a ``sunny day'' even if the sunlight appears diffused. Conversely, if the road surface appears wet or snow-covered—even in the absence of ongoing precipitation—we annotate the condition as ``rainy'' or ``snowy'' due to the increased risk of slipping and reduced friction. These decisions are made from a safety-critical perspective, prioritizing conditions that affect vehicle control or pedestrian stability. More detailed visual cues regarding surface wetness, lighting, or reflections are included in the dense caption descriptions.

\paragraph{Traffic Environment}
Accidents involving VRUs generally occur in either urban or rural settings. The environmental classification is relatively straightforward based on visual cues such as infrastructure density, type of roadside elements, and presence of intersections or buildings. In most cases, the classification is unambiguous.

\paragraph{Road Configuration}
Although many videos provide a clear view of the road structure, the moving dashcam perspective can sometimes obscure precise location categorization. For example, if a vehicle is approaching an intersection, the same video segment may contain both arterial roads and the intersection itself. In such cases, we annotate the configuration based on where the key event (e.g., collision or near-miss) occurs. If the impact takes place just before entering the intersection, the label is set to ``arterial road"; if the event occurs within the crossroad, it is labeled as ``intersection." The sequential progression and transitions in road context are described in detail within the dense captions.

\paragraph{Accident Type}
While most accident types can be clearly and concisely described (e.g., “car hits pedestrian,” “bicycle falls”), there are borderline or ambiguous cases. For instance, when a dashcam-equipped vehicle rapidly approaches a pedestrian and the pedestrian narrowly avoids contact by sidestepping, it becomes unclear whether to classify the event as an actual collision or a near-miss. In such cases, we annotate based on the interaction and directionality of the entities involved. If a pedestrian avoids a moving ego-vehicle by inches, the event is still annotated as “ego-car hits a pedestrian” for the sake of interpretability and consistency. Similarly, if a cyclist accidentes into a parked vehicle, it is labeled as “car hits cyclist,” following common conventions seen in prior datasets such as \cite{MM_AU, CAP_DATA}. These nuances are captured more fully in the dense captions, where the dynamic unfolding of the scenario is described.

\paragraph{Accident Cause}
Causality in traffic accidents is often multi-factorial and temporally distributed. It is difficult to capture such complexity within a short answer choice. Therefore, our annotations focus on the most immediate and visually interpretable cause of the accident—such as driver inattention, sudden lane change, or pedestrian jaywalking. These labels are selected based on high-salience cues that directly precede the event. The dense caption annotations, however, elaborate on the sequence of road user actions, delays in reaction, or misjudgments over time that contributed to the accident.

\paragraph{Accident Prevention Measure}
This category is annotated based on what preventive action could have reasonably avoided the accident, considering the behavior of all involved road users. Rather than speculating on counterfactuals involving unrealistic interventions, we focus on plausible, context-aware measures that align with traffic norms and human capabilities. For example, if a vehicle accelerated despite a pedestrian entering the crosswalk, the prevention measure might be “driver should yield to pedestrian.” In another case, if a cyclist riding without a helmet crosses the road illegally during rainy conditions and collides with a vehicle, the core cause of the accident is the illegal crossing. Among multiple plausible prevention strategies—such as ``cyclist should wear a helmet,'' ``cyclist should wear bright clothing,'' ``cyclist should reduce speed,'' and ``cyclist should use designated crosswalk''—we annotate the most direct and effective preventive measure, in this case: ``cyclist should use designated crosswalk.'' This enables our benchmark to assess whether models can move beyond reliance on visually salient but potentially misleading cues, and instead demonstrate high-level understanding by identifying the true cause of the accident and articulating the most appropriate preventive action. In other words, it evaluates the model’s ability to reason about causality in safety-critical scenarios rather than simply describing visually correct but contextually irrelevant information.

\begin{figure}[t!]
\includegraphics[width=\columnwidth]{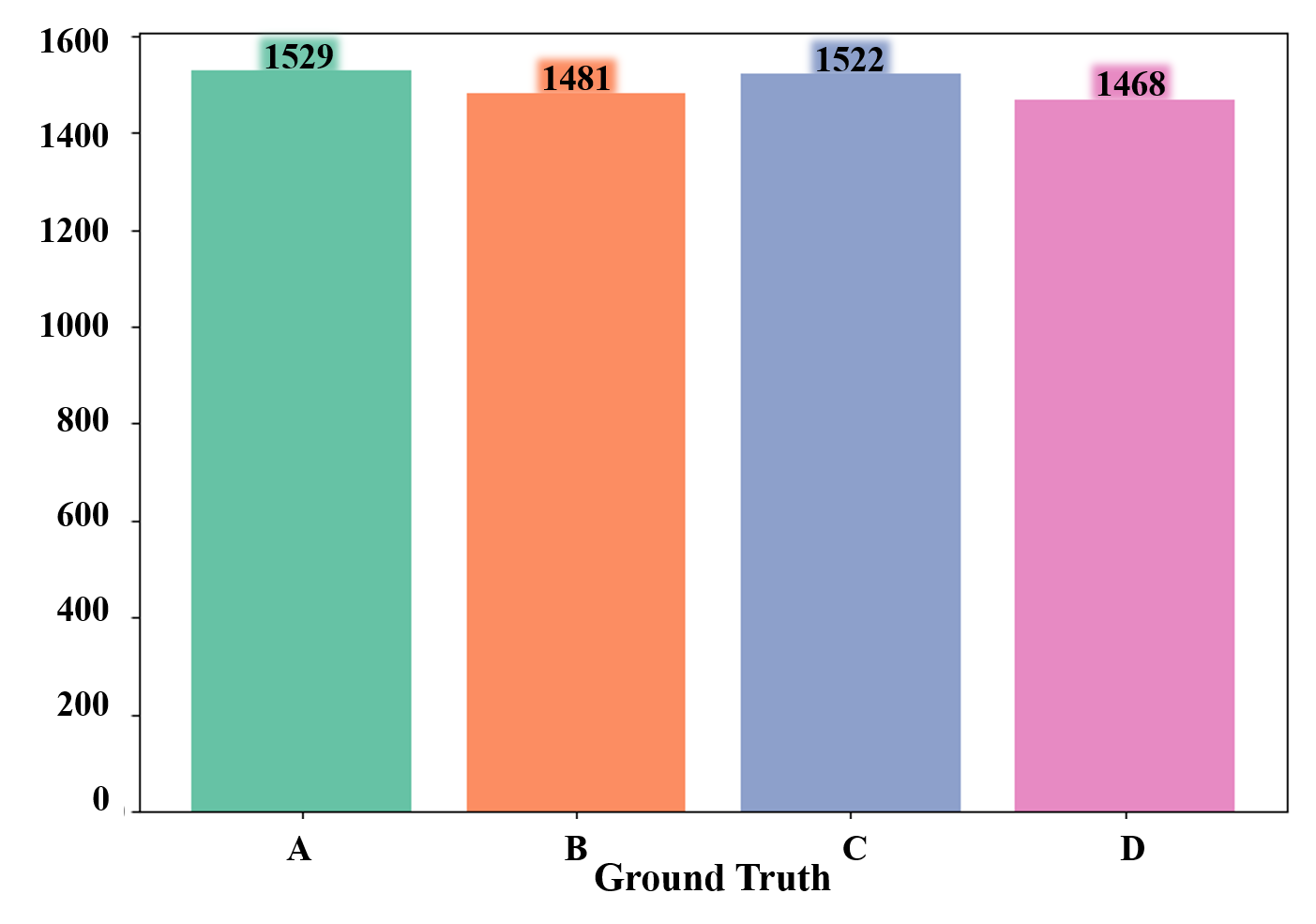}
\centering
\caption{Distribution of ground truth of the VQA task in the VRU-Accident benchmark.}
\label{fig:VQA_GT_distribution}
\end{figure}

\subsection{VQA Statistics}

\paragraph{Word Distribution per Category}  
As illustrated in Figure~\ref{fig:VQA_word_distribution}, the answer options across six VQA categories exhibit a wide range of lexical diversity. Each category---ranging from \textit{Weather \& Light Condition} to \textit{Accident Prevention Measure}---contains semantically rich and diverse phrases. This diversity increases the difficulty of answer selection, requiring high-level scene understanding and contextual reasoning beyond keyword matching. Therefore, the benchmark provides a suitable evaluation setting for probing the generalization capability and multimodal grounding ability of MLLMs.

\paragraph{Balanced Ground Truth Distribution}  
Figure~\ref{fig:VQA_GT_distribution} presents the overall distribution of ground truth answers (A–D) in the VRU-Accident benchmark. The distribution remains relatively balanced across all four options, with less than 5\% deviation between the most and least frequent labels. This balance mitigates bias toward any specific answer index and ensures reliable evaluation without prior answer frequency bias, making the benchmark statistically robust for multi-choice classification tasks.

\paragraph{Semantic Similarity Between Question and Answers}  
To assess the semantic relevance between questions and answer options in VRU-Accident, we compute the cosine similarity between their embeddings using the \texttt{all-MiniLM-L6-v2} model~\cite{Similarity}. We denote the embedding model as a function \( F(\cdot) \), which maps each sentence into a fixed-dimensional vector. For a given question \( q \) and its four answer options \( a_1, a_2, a_3, a_4 \), the cosine similarity between the question and each option is calculated as:

\begin{equation}
\text{sim}(q, a_k) = \frac{\langle F(q), F(a_k) \rangle}{\|F(q)\| \cdot \|F(a_k)\|},
\end{equation}

where \( k \in \{1,2,3,4\} \). For each option \( k \), we evaluate the average similarity of category \(j\) between the question and the ground truth answer over all \( N \) samples. This average ground truth similarity is computed as:

\begin{equation}
\text{Sim}^{\text{GT}}_j = \frac{1}{N} \sum_{i=1}^{N} \text{sim}(q^{(i)}, a_{g^{(i)}}^{(i)}),
\end{equation}

where \( a_{g^{(i)}}^{(i)} \) is the ground truth answer for the \( i \)-th sample. To assess the relevance of distractors, we take the three incorrect options for each sample, compute their similarity to the question, and average them across the dataset:

\begin{equation}
\text{Sim}^{\text{Dist}}_j = \frac{1}{3N} \sum_{i=1}^{N} \sum_{\substack{k=1 \\ k \neq g^{(i)}}}^{4} \text{sim}(q^{(i)}, a_k^{(i)}).
\end{equation}

\begin{figure}[t!]
\includegraphics[width=\columnwidth]{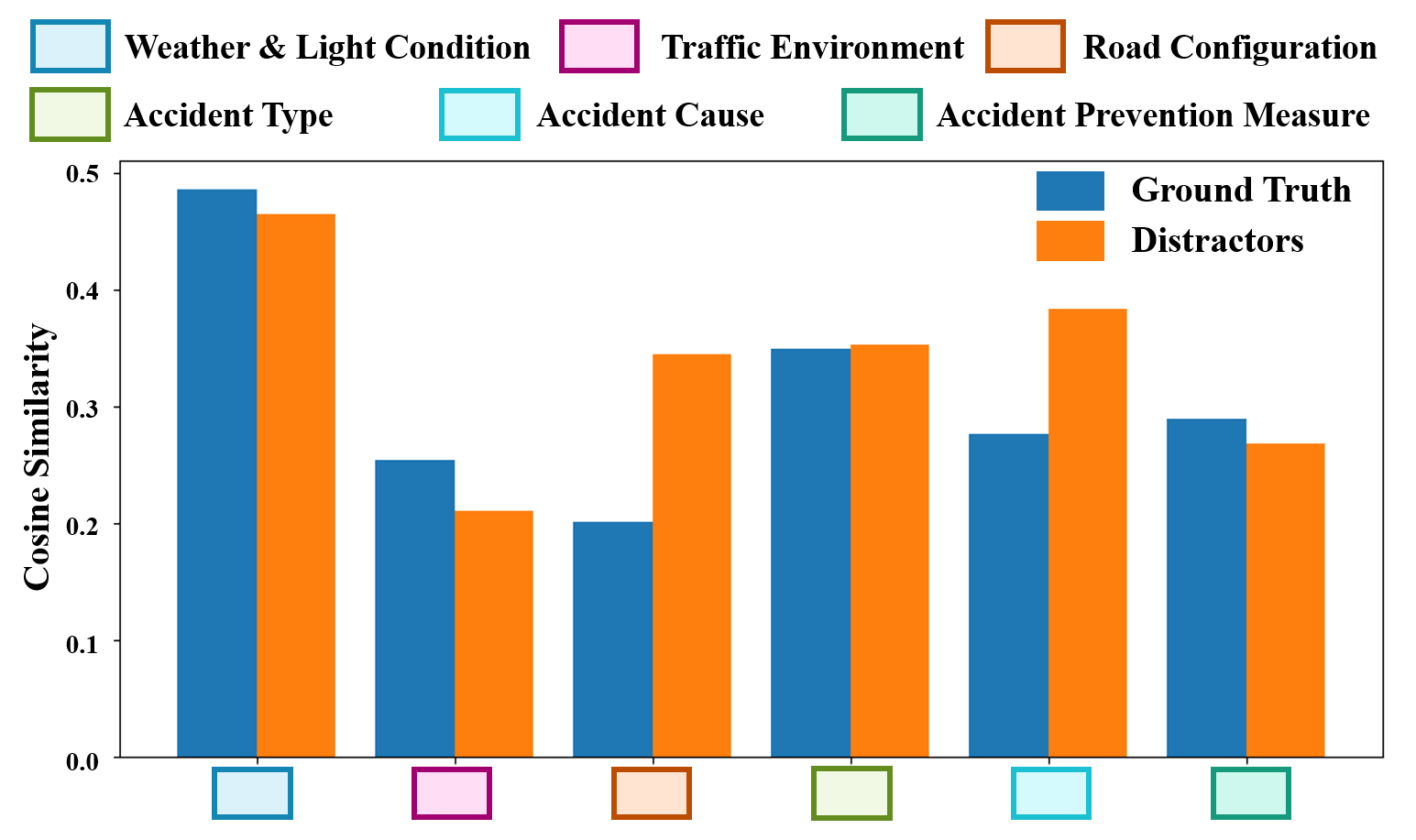}
\centering
\caption{Comparison of average Q-A cosine similarity for ground truth answers and distractors across VQA categories in VRU-Accident benchmark.}
\label{fig:VQA_QA_similarity}
\end{figure}

Figure~\ref{fig:VQA_QA_similarity} shows that distractors in several categories exhibit comparable or even higher similarity to the question than the ground truth. This demonstrates that the distractor options are semantically plausible and not easily distinguishable without detailed reasoning. Therefore, this benchmark poses a challenging evaluation setting for assessing the semantic discrimination and contextual understanding capabilities of multimodal large language models (MLLMs).

\section{Qualitative Examples}
\label{sec:example}

\begin{figure*}[!t]
    \centerline{\includegraphics[width=\textwidth]{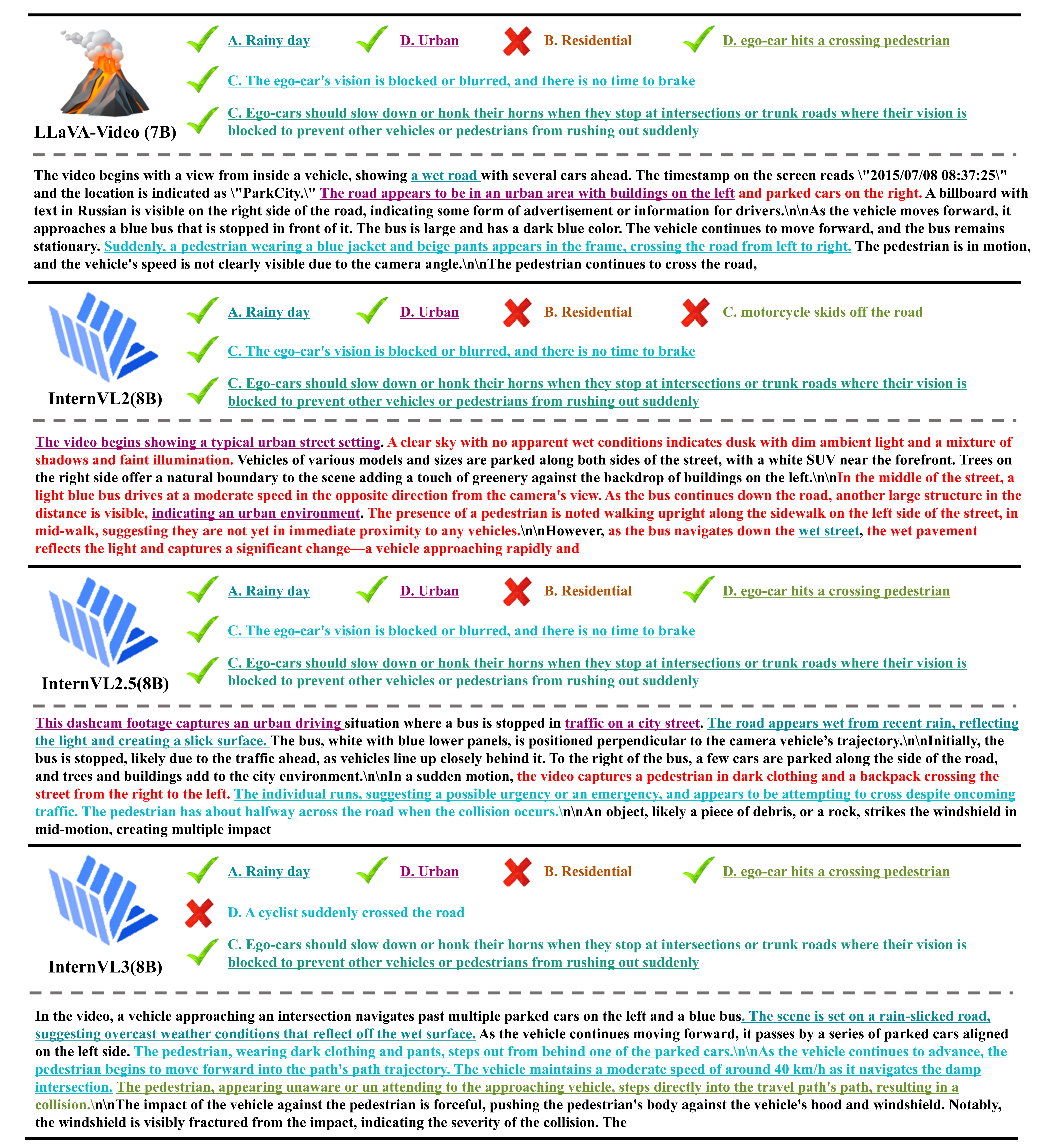}}
    \caption{Qualitative comparison of VQA and dense captioning outputs from MLLMs on the VRU-Accident benchmark. Each example shows the ground truth annotation (top row) and the responses from models.}
    \label{fig:Example_Visualization}
\end{figure*}

To further analyze model performance, we present qualitative comparisons between the ground truth annotations in VRU-Accident and the responses generated by state-of-the-art MLLMs~\cite{LLaVA_OneVision, Mobile_VideoGPT, InternVL25, InternVL3, InternVL2, qwen2_VL, qwen25_VL, Video_XL_Pro, Video_XL2, LLaVA_NeXT_Video, LLaVA_Video} for both the VQA and dense captioning tasks. Figure~\ref{fig:Example_Visualization} illustrates a representative set of cases, offering insight into the models' capabilities and limitations in understanding complex accident scenarios.

We observe that many models correctly identify basic visual attributes (e.g., rainy day or urban setting). However, several limitations remain evident. For instance, while dense captions often include fluent descriptions of the scene layout and pedestrian appearances, models frequently hallucinate motion trajectories or misattribute collision responsibility, especially in complex scenarios involving occlusion or sudden movements.

\begin{figure*}[!t]
    \centerline{\includegraphics[width=\textwidth]{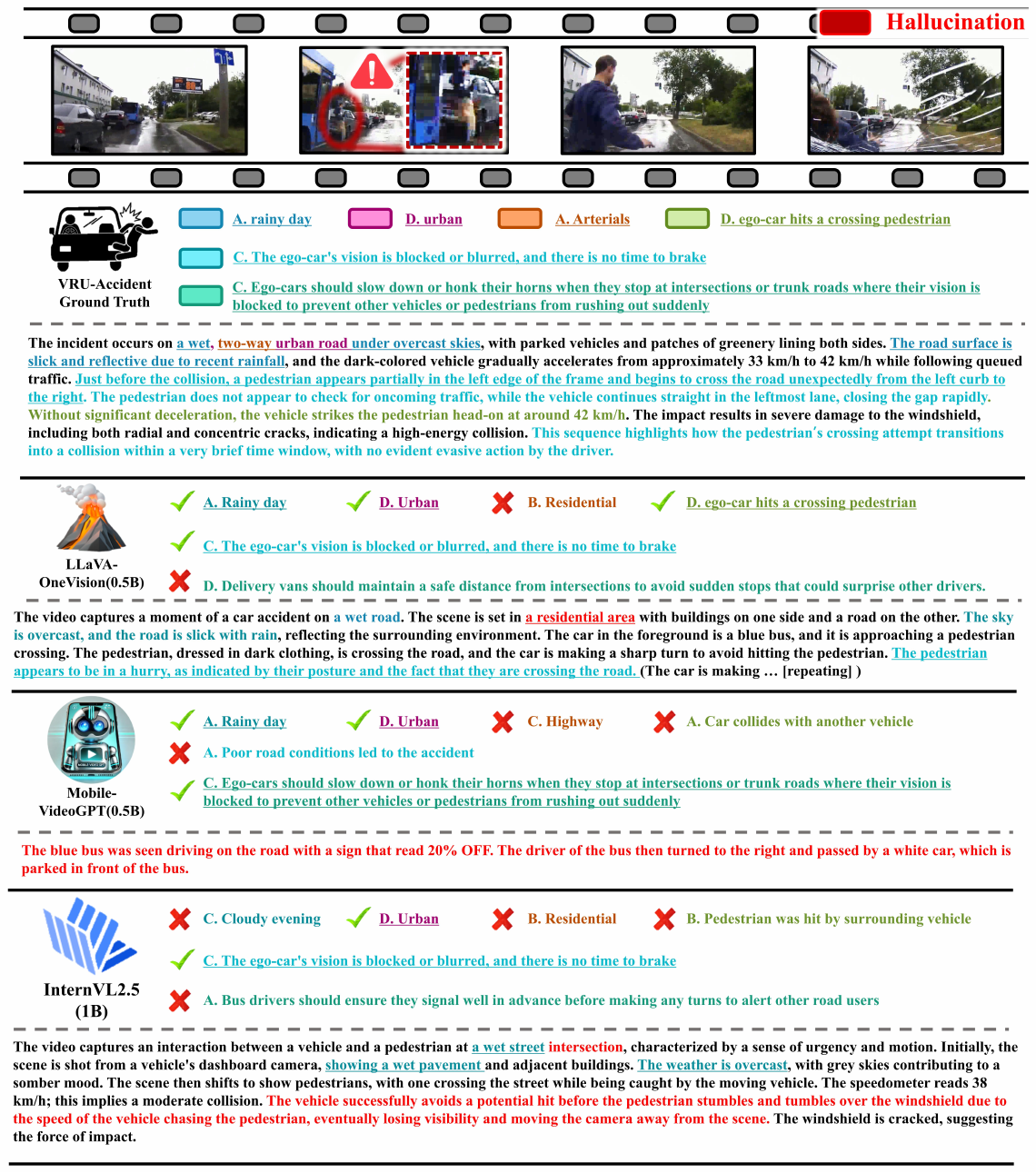}}
\end{figure*}

\begin{figure*}[!t]
    \centerline{\includegraphics[width=\textwidth]{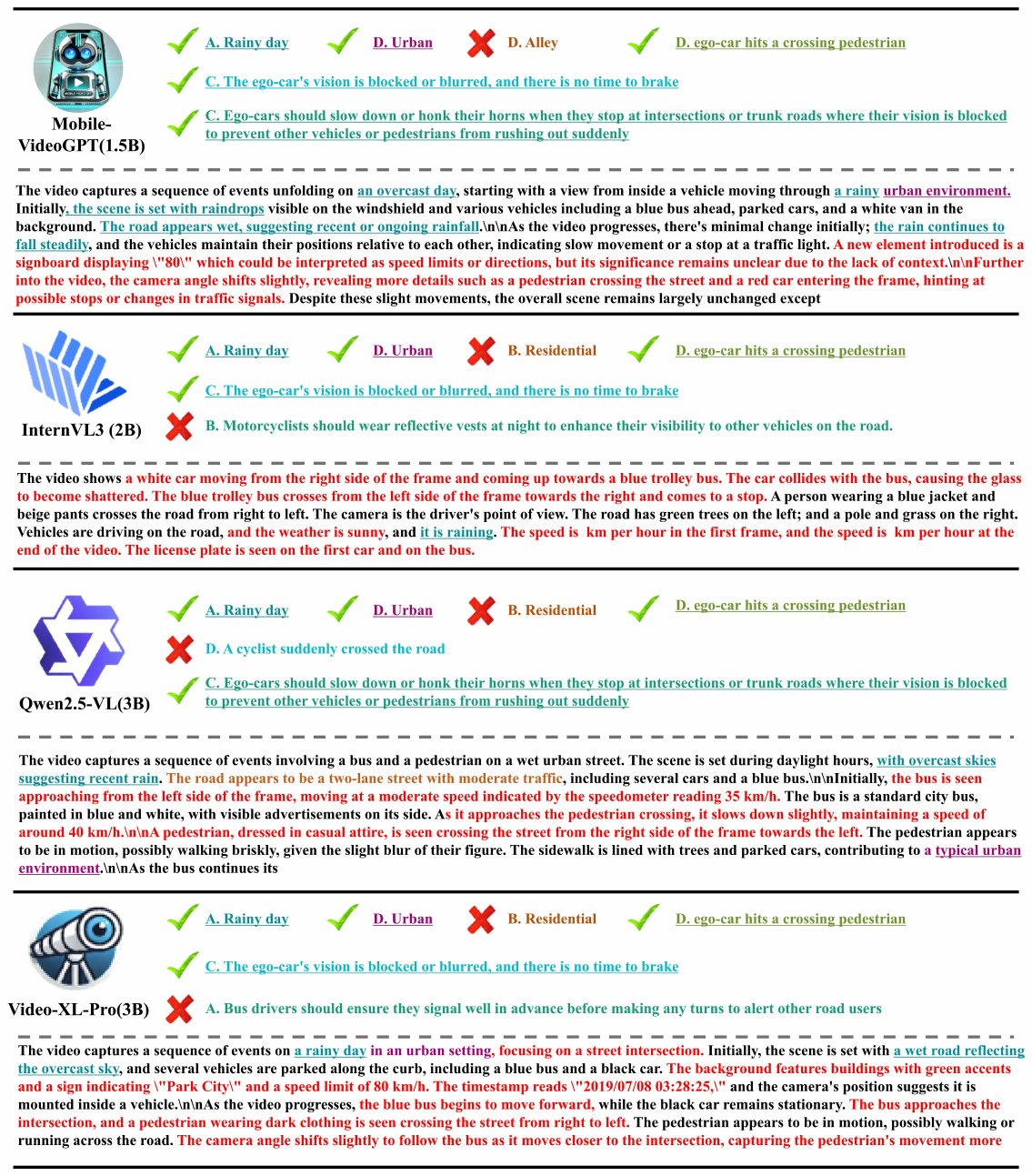}}
\end{figure*}

\begin{figure*}[!t]
    \centerline{\includegraphics[width=\textwidth]{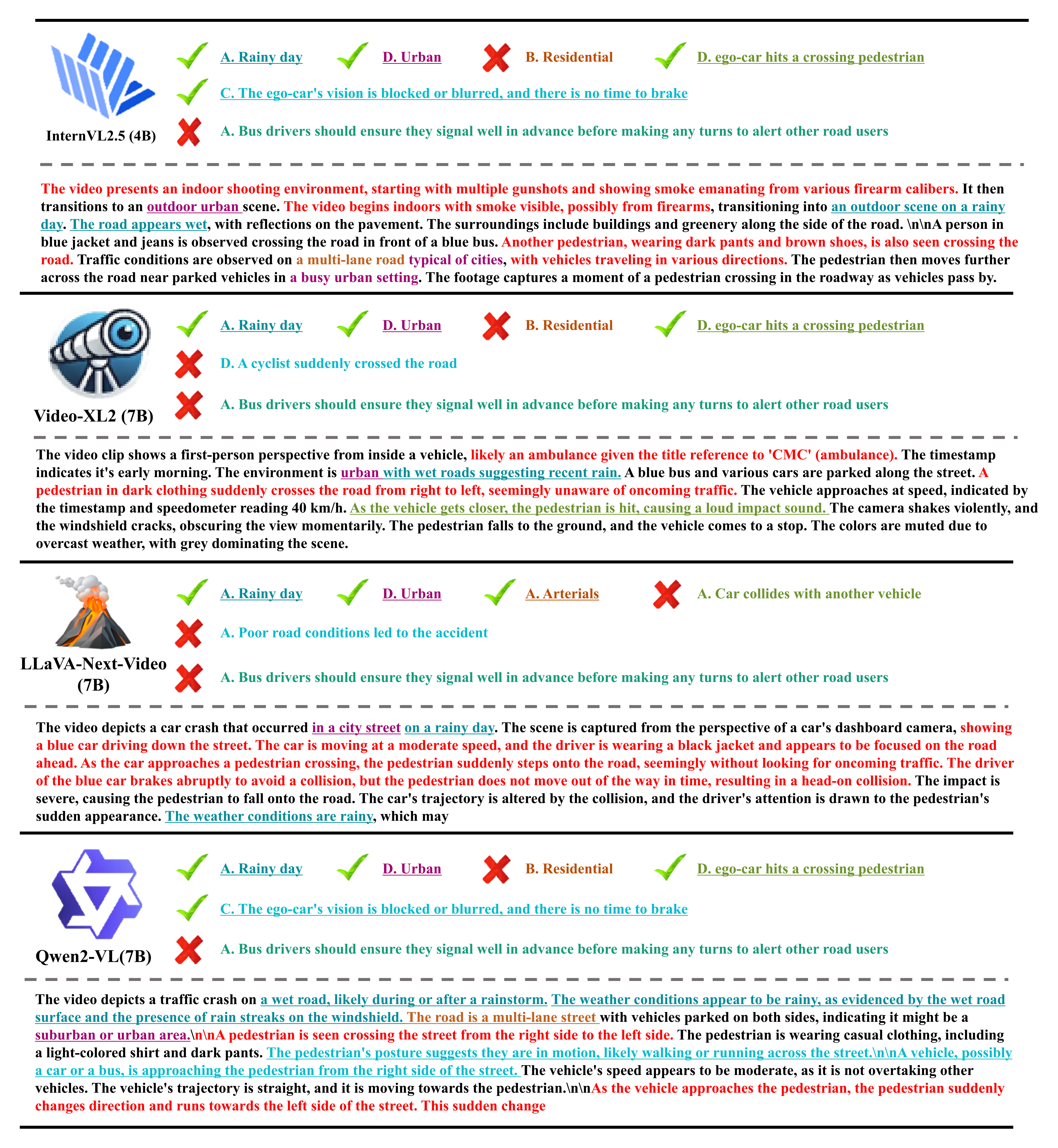}}
\end{figure*}

These examples underscore the value of VRU-Accident as a challenging benchmark. It not only requires visual recognition but also demands high-level temporal and causal reasoning. By contrasting ground truth annotations with model predictions, our qualitative analysis reveals both the current capabilities and the remaining gaps in modern MLLMs’ ability to comprehend and explain real-world accidents involving vulnerable road users.

\section{Prompts for VRU-Accident Curation}
\label{sec:prompt}

\begin{figure*}[!t]
    \centerline{\includegraphics[width=\textwidth]{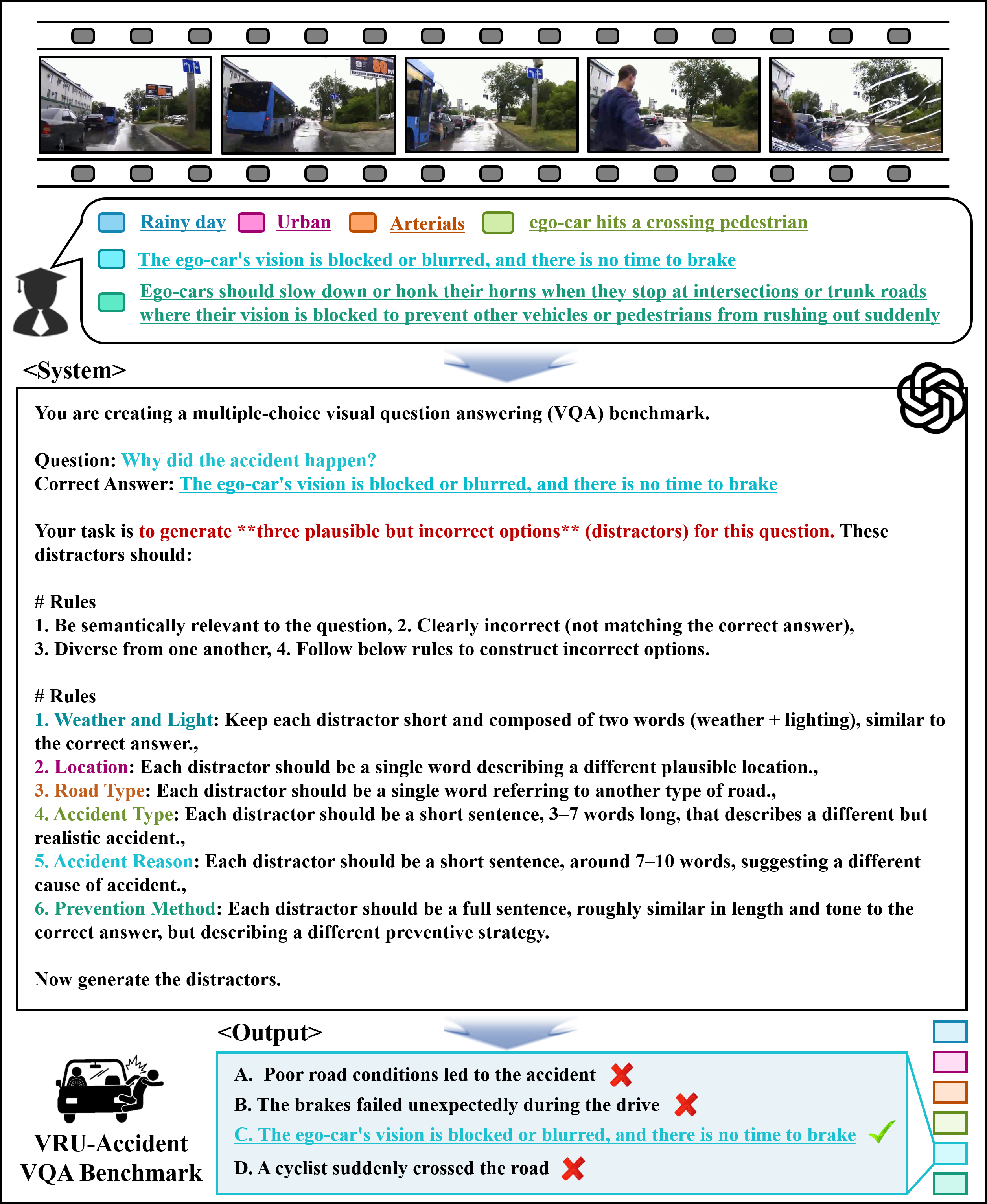}}
    \caption{Prompt used for curating VRU-Accident VQA benchmark, including category-specific rules and a generated multiple-choice question with one correct answer and three semantically plausible distractors.}
    \label{fig:VQAPrompt}
\end{figure*}

\paragraph{Prompts for VQA Curation}
Figure~\ref{fig:VQAPrompt} illustrates the detailed prompt used to curate the VQA benchmark in VRU-Accident. Human annotators first reviewed each event video and manually labeled the correct answer for each reasoning category. Based on this ground truth and the corresponding question, GPT-4o was employed to generate three plausible but incorrect options (distractors) that preserve semantic relevance but differ from the correct answer. The example in Figure~\ref{fig:VQAPrompt} involves a rainy-day accident where a pedestrian suddenly emerges from between parked vehicles, resulting in a collision. For the \textbf{Accident Reason} category, the ground truth is “C. The ego-car's vision is blocked or blurred, and there is no time to brake.” GPT-4o generates the following distractors: “A. Poor road conditions led to the accident,” “B. The brakes failed unexpectedly during the drive,” and “D. A cyclist suddenly crossed the road.” Option A, while weather-related, is incorrect because the road condition was not the direct cause. Option B is wrong as the brakes were functioning, and D is incorrect since the involved road user was a pedestrian, not a cyclist. These semantically plausible yet incorrect choices help assess whether MLLMs can go beyond surface-level cues and demonstrate high-level causal reasoning.

\begin{figure*}[!t]
    \centerline{\includegraphics[width=\textwidth]{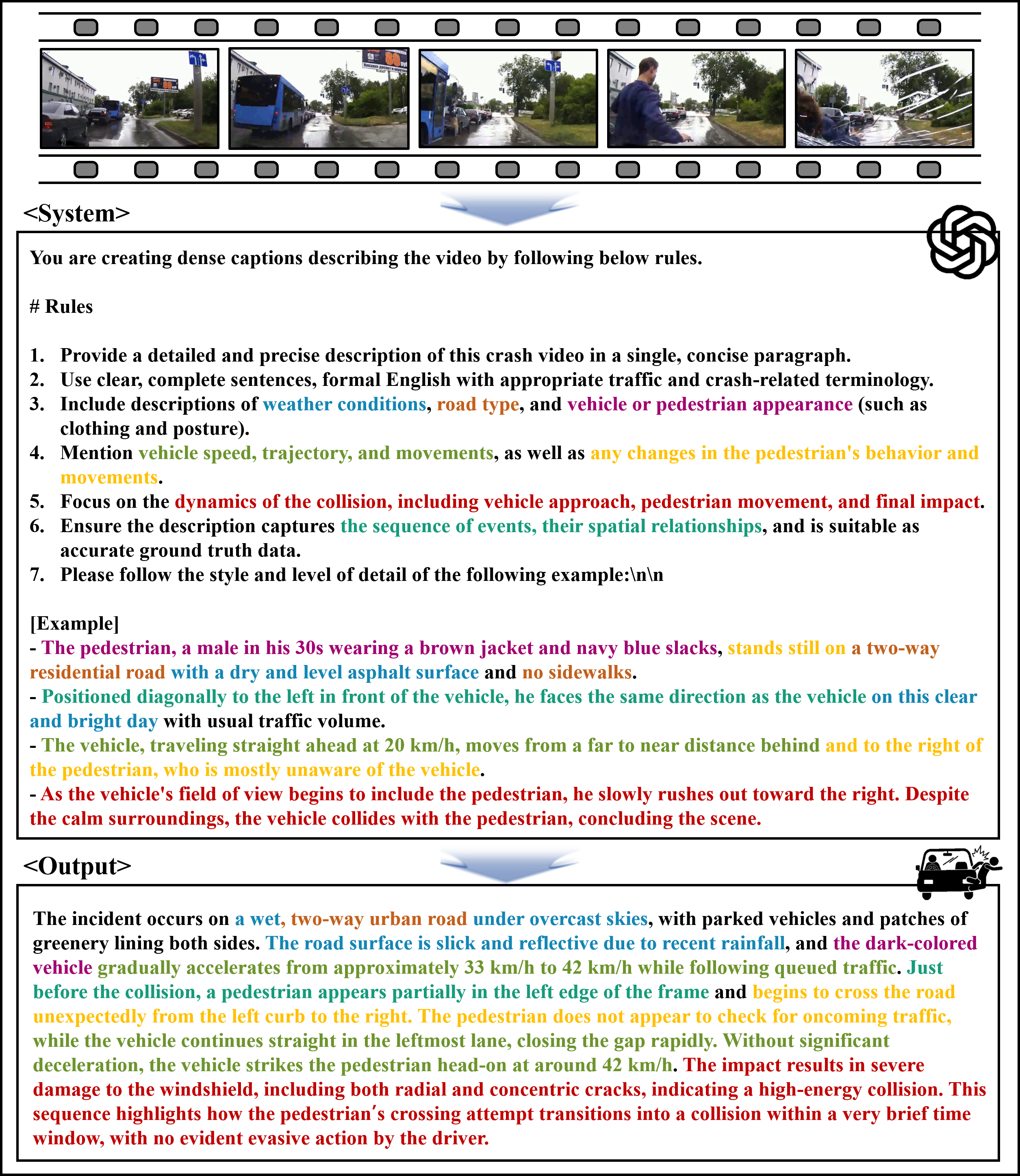}}
    \caption{Prompt used for curating dense caption annotations in VRU-Accident. The prompt guides the model to describe accident videos with detailed references to weather, road configuration, agent appearance, kinematic features, and collision dynamics.}
    \label{fig:DenseCaptionPrompt}
\end{figure*}

\paragraph{Prompts for Dense Caption Curation}
Figure~\ref{fig:DenseCaptionPrompt} shows the prompt used to curate dense accident descriptions. The prompt instructs the model to describe each accident video with detailed and coherent narratives, covering weather conditions, road type, vehicle and pedestrian appearances, vehicle dynamics (e.g., speed and trajectory), pedestrian behavior, spatial and temporal relationships among road users, and final collision outcomes. GPT-4o takes the input video and prompt rules to generate a single paragraph that mirrors the descriptive style and level of detail found in human-written annotations. This design ensures that the generated captions are not only visually grounded but also demonstrate temporal reasoning and contextual understanding of the accident dynamics. As a result, the VRU-Accident benchmark enables qualitative assessment of MLLMs’ ability to comprehend and articulate complex accident scenarios beyond mere visual recognition.

\section{Reproduction of Experiment}
\label{sec:reproduction}

To encourage widespread use of our VRU-Accident benchmark and to support reproducibility of our experiments, we release all necessary resources used for all experiments in the main manuscript. This includes the inference scripts, model outputs, and evaluation results. Please visit our Github\footnote{\url{https://github.com/Kimyounggun99/VRU-Accident}} and refer the \textbf{ReadMe} file for reproduction.

\section{Acknowledgements}
\label{sec:acknowledgements}
We would like to express sincere gratitude to Uibeom Chun (University of Central Florida) for his invaluable assistance with the conceptualization and experiments in this work. His contributions greatly supported the development of this research.

\newpage


\end{document}